\documentclass[journal]{IEEEtran}

\usepackage{graphicx}

\usepackage{algorithm}
\usepackage{amsmath}
\usepackage{calligra}
\DeclareMathAlphabet{\mathcalligra}{T1}{calligra}{m}{n}

\usepackage{array}
\usepackage{enumerate}
\usepackage{multicol}
\usepackage{multirow}
\usepackage{mathtools}
\usepackage{stmaryrd}

\usepackage{mathptmx} 
\usepackage{threeparttable}
\usepackage{algpseudocode}
\usepackage{cite}
\usepackage{array}
\usepackage{xcolor, colortbl}
\definecolor{Gray}{gray}{0.9}
\definecolor{LightCyan}{rgb}{1,1,1}
\usepackage{amssymb}
\usepackage{amsfonts}
\usepackage{amsmath}
\usepackage{amsfonts}
\usepackage{verbatim}
\usepackage{tasks}
\usepackage{tcolorbox}
\usepackage{lineno}
\usepackage[utf8]{inputenc}
\usepackage[english]{babel}

\usepackage{booktabs}
\PassOptionsToPackage{hyphens}{url}
\usepackage{hyperref}

\usepackage{subcaption, microtype}

\DeclareMathAlphabet{\mathpzc}{OT1}{pzc}{m}{it}
\DeclareMathAlphabet\mathbfcal{OMS}{cmsy}{b}{n}

\makeatletter
\newcommand{\xMapsto}[2][]{\ext@arrow 0599{\Mapstofill@}{#1}{#2}}
\def\Mapstofill@{\arrowfill@{\Mapstochar\Relbar}\Relbar\Rightarrow}

\modulolinenumbers[5]

\begin{document}

%
\title{A Generic Approach to Lung Field Segmentation from Chest Radiographs using Deep Space and Shape Learning}
%
%
%

\author{Awais~Mansoor\textsuperscript{*},
        Juan~J.~Cerrolaza,
				Geovanny~Perez,
				Elijah~Biggs,
				Kazunori Okada,
				Gustavo~Nino,\\
        Marius~George~Linguraru.
\thanks{\textit{ Asterisk indicates corresponding author:} awais.mansoor@gmail.com.}
\thanks{A. Mansoor, J. J. Cerrolaza, and E. Biggs are with the Sheikh Zayed Institute for Pediatric Surgical Innovation, Children's National Health System, Washington DC.}
\thanks{G. Perez and G. Nino are with the Division of Pulmonary and Sleep Medicine, Children’s National Health System, Washington, DC.}
\thanks{K. Okada is with Computer Science Department, San Francisco State University, San Francisco, CA.}
\thanks{M. G. Linguraru is with the Sheikh Zayed Institute for Pediatric Surgical Innovation, Children's National Health System and the School of Medicine and Health Sciences, George Washington University, Washington DC.}
}

\maketitle

\begin{abstract}
Computer-aided diagnosis (CAD) techniques for lung field segmentation from chest radiographs (CXR) have been proposed for adult cohorts, but rarely for pediatric subjects. Statistical shape models (SSMs), the workhorse of most state-of-the-art CXR-based lung field segmentation methods, do not efficiently accommodate shape variation of the lung field during the pediatric developmental stages. The main contributions of our work are: (1) a generic lung field segmentation framework from CXR accommodating large shape variation for adult and pediatric cohorts; (2) a deep representation learning detection mechanism, \emph{ensemble space learning}, for robust object localization; and (3) \emph{marginal shape deep learning} for the shape deformation parameter estimation. Unlike the iterative approach of conventional SSMs, the proposed shape learning mechanism transforms the parameter space into marginal subspaces that are solvable efficiently using the recursive representation learning mechanism. Furthermore, our method is the first to include the challenging retro-cardiac region in the CXR-based lung segmentation for accurate lung capacity estimation. The framework is evaluated on 668 CXRs of patients between 3 month to 89 year of age. We obtain a mean Dice similarity coefficient of $0.96\pm0.03$ (including the retro-cardiac region). For a given accuracy, the proposed approach is also found to be faster than conventional SSM-based iterative segmentation methods. The computational simplicity of the proposed generic framework could be similarly applied to the fast segmentation of other deformable objects.
\end{abstract}


\IEEEpeerreviewmaketitle

%
%
%
%
\section{Introduction}
\begin{figure}[htb]
\begin{subfigure}[b]{0.155\textwidth}
\includegraphics[width=\textwidth]{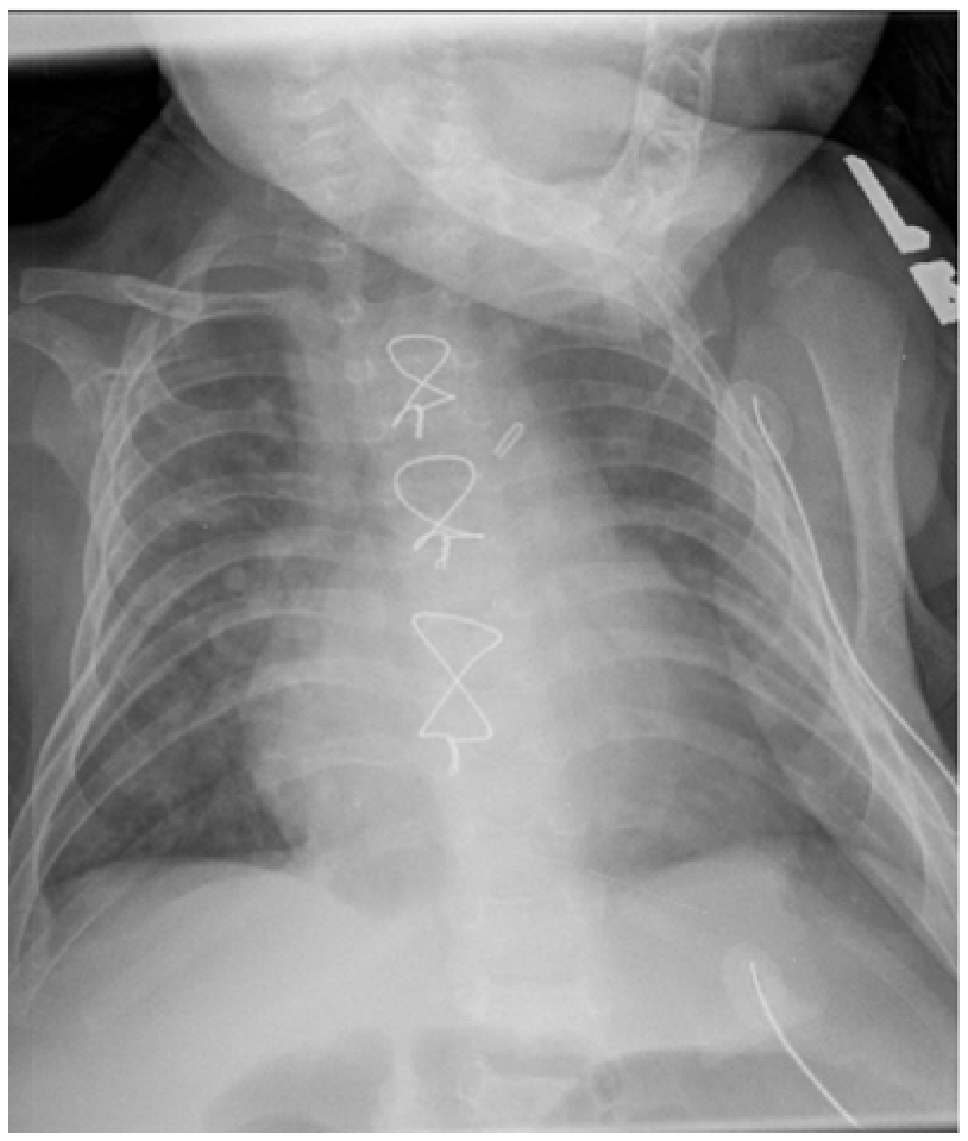}
\caption{}
\end{subfigure}
\begin{subfigure}[b]{0.155\textwidth}
\includegraphics[width=\textwidth]{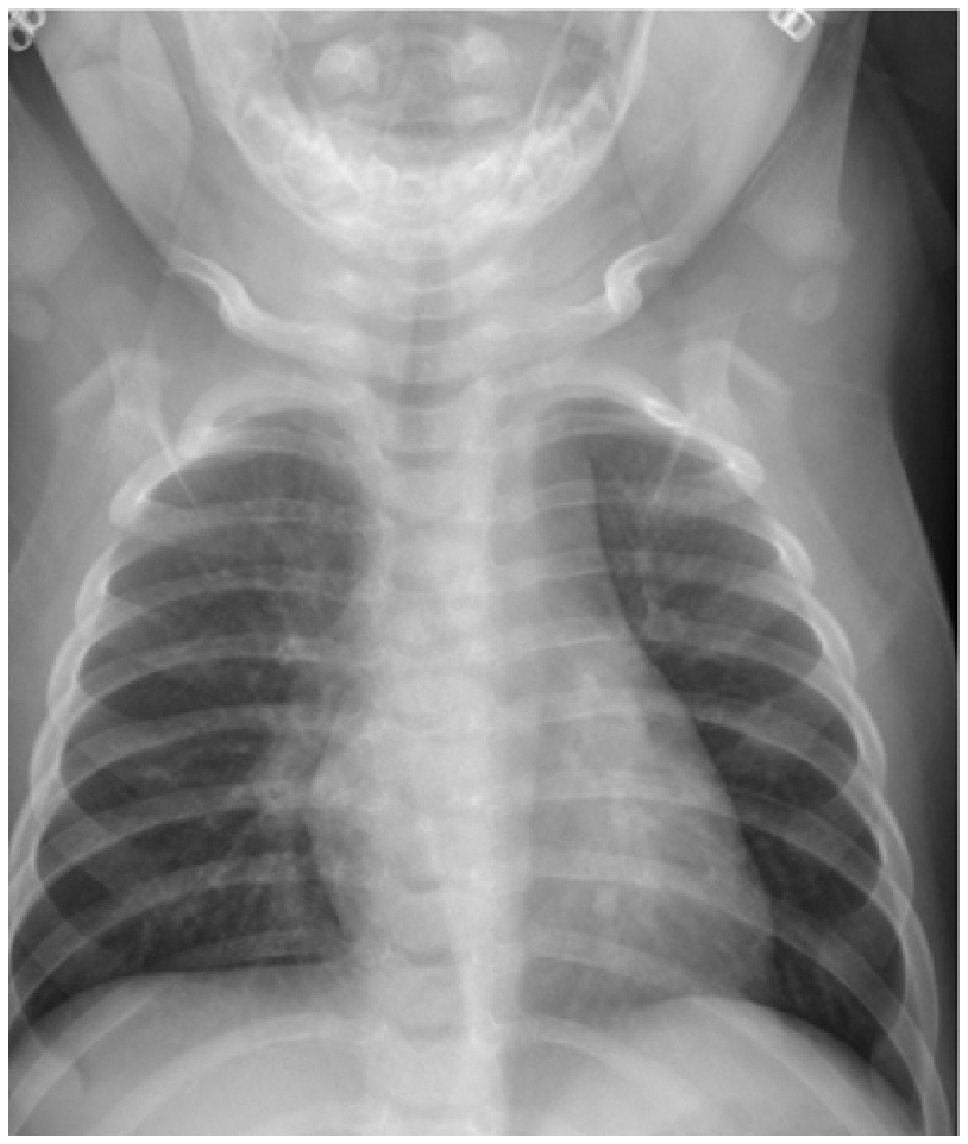}
\caption{}
\end{subfigure}
\begin{subfigure}[b]{0.155\textwidth}
\includegraphics[width=\textwidth]{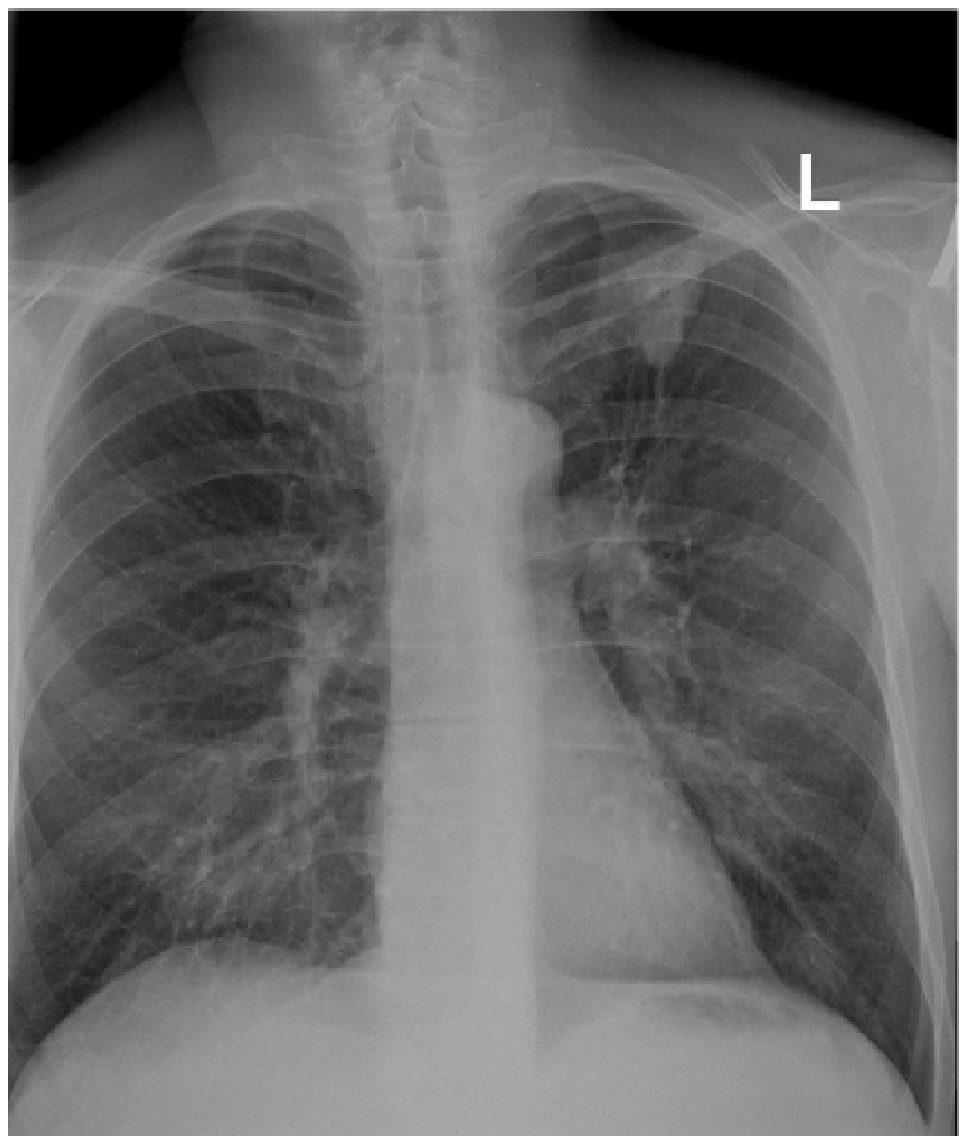}
\caption{}
\end{subfigure}
\begin{subfigure}[b]{0.24\textwidth}
\includegraphics[width=\textwidth]{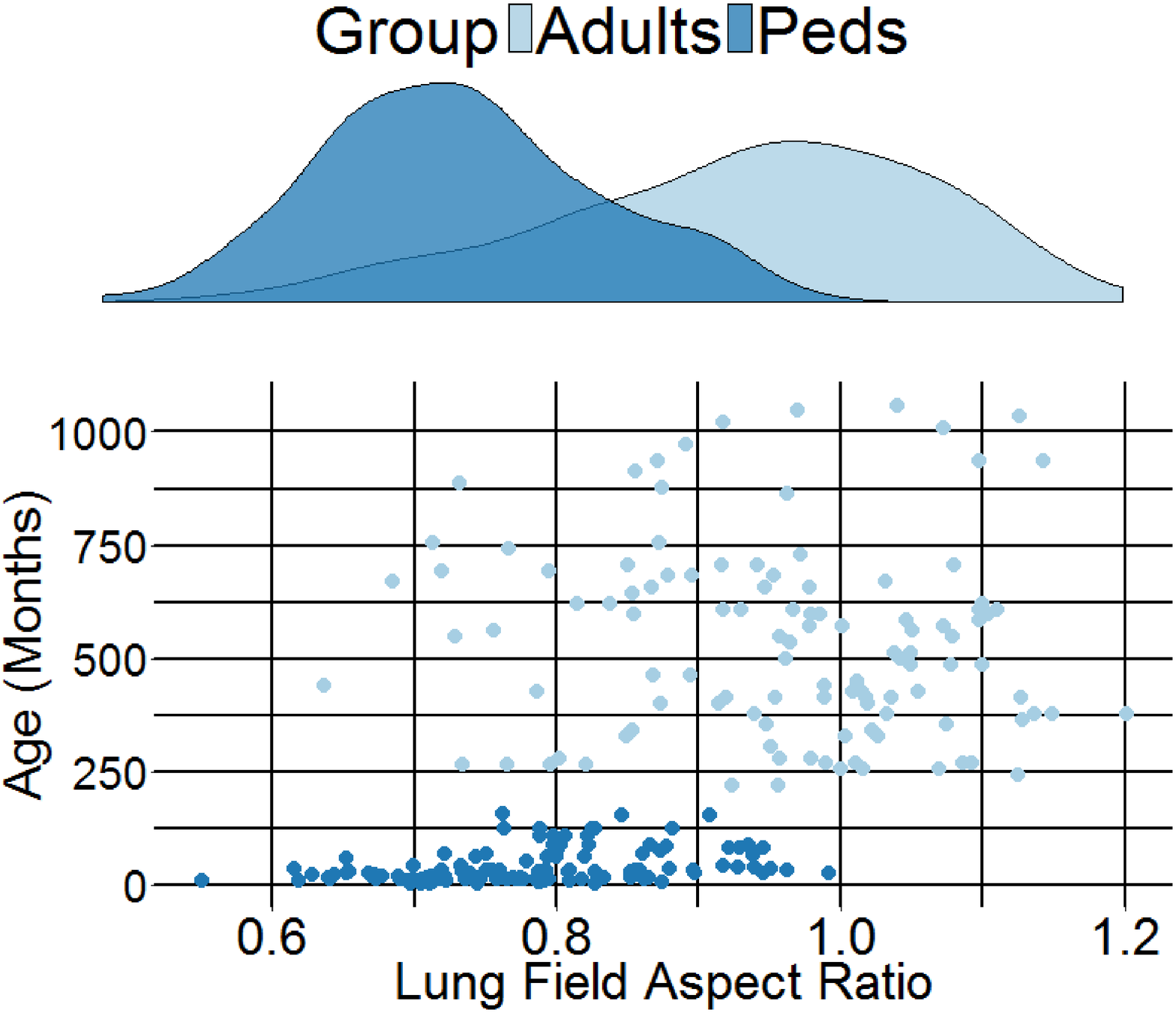}
\caption{}
\label{fig:adultPedsAspectRatio}
\end{subfigure}
\begin{subfigure}[b]{0.24\textwidth}
\includegraphics[width=\textwidth]{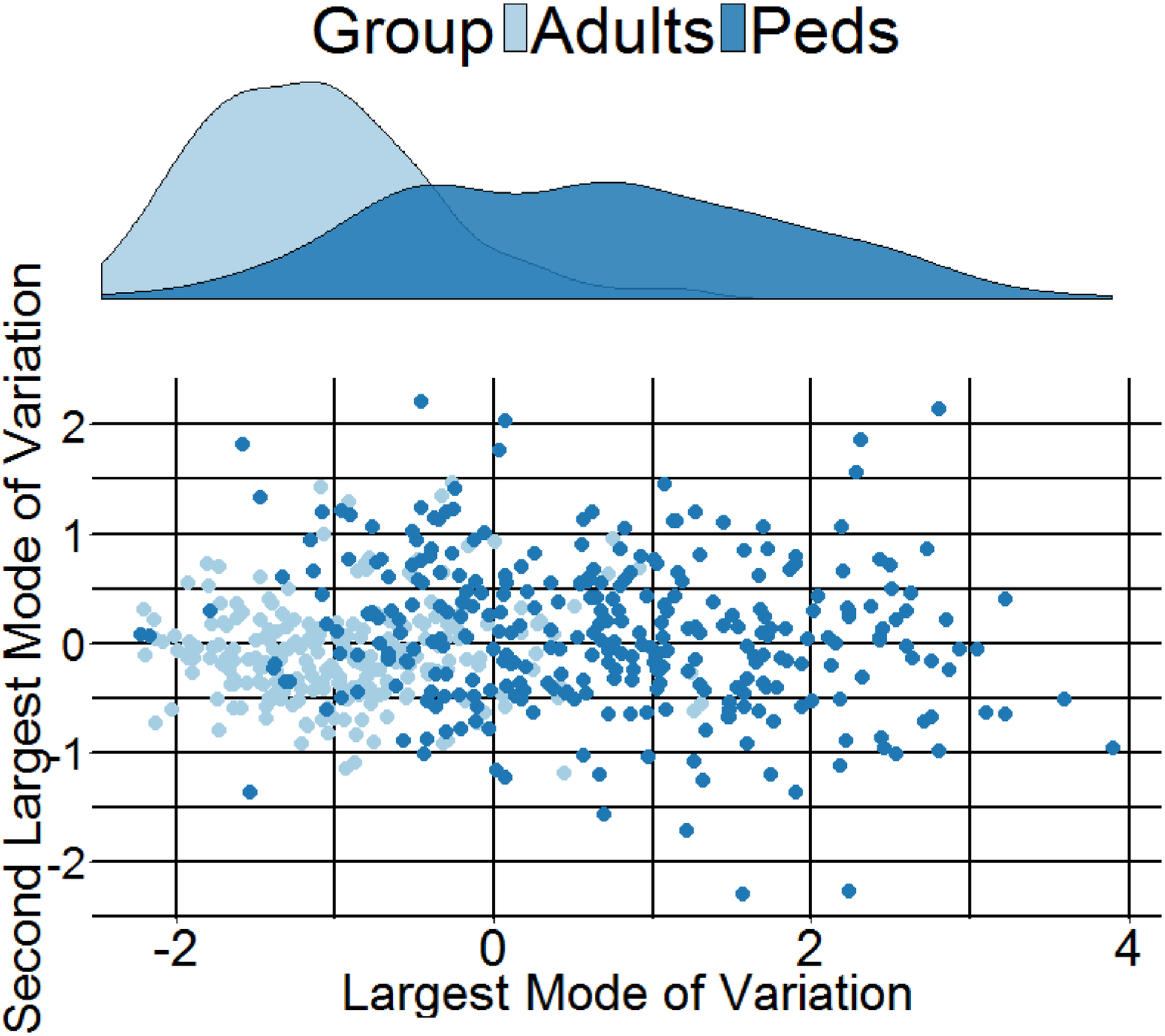}
\caption{}
\label{fig:adultPedsPCA}
\end{subfigure}
\caption{\footnotesize{Illustration of age-related anatomical differences captured within CXRs. CXR obtained from: (a) 2-month old subject, (b) 4-year old subject, (c) 44-year old subject. (d) Structural differences in the lung field between the adults and pediatrics based on the aspect ratio, (e) Structural differences in the lung field between the adults and pediatrics based on the two largest modes of principal component analysis.}}
\label{fig:adultPeds}
\end{figure}

Despite tremendous advancements in tomographic imaging, chest radiography remains the most commonly used imaging modality for pulmonary analysis mainly due to its low cost, low radiation dosage, and widespread availability. Radiation dosage is of particular concern in pediatric applications, especially in neonatal intensive care units where chest radiographs (CXRs) are considered the first option for pulmonary diagnosis \cite{yu2010radiation}. Lung field segmentation is the necessary initial step for image-based pulmonary analysis. Accurate delineation of lung field from CXR, however, is challenging due to ambiguous boundaries, pathologies, occultation of lung field by anatomical structures in thorax, anatomical variation of lung shapes, and size across subjects (Fig. \ref{fig:adultPeds}). Part of the challenge in developing computer-aided diagnosis (CAD) methods, especially for pediatric cohorts, is also the anatomical shape variation of lung field that occur during growth \cite{candemir2015lung, smeets1990lung}. As shown in Fig. \ref{fig:adultPeds}, pediatric cohorts have a more compliant chest wall, small thoracic cage, and relative large abdominal space. Furthermore, the diaphragm of children has smaller apposition area which has a concave shape in the posterior-anterior (PA) view CXR \cite{smeets1990lung}. Therefore, existing approaches to lung field segmentation that are designed primarily for adult cohorts, are not accurate at analyzing the pediatric subjects. Although a few pilot studies such as \cite{candemir2015lung} have been conducted recently to look at the age-related radiological biomarkers in lungs, no comprehensive study of pediatric lung field segmentation exists to the best of our knowledge.

\begin{figure}
\begin{minipage}{.15\textwidth}
\begin{minipage}{\textwidth}
\begin{subfigure}[c]{0.9\textwidth}
\centering
\includegraphics[width=\textwidth]{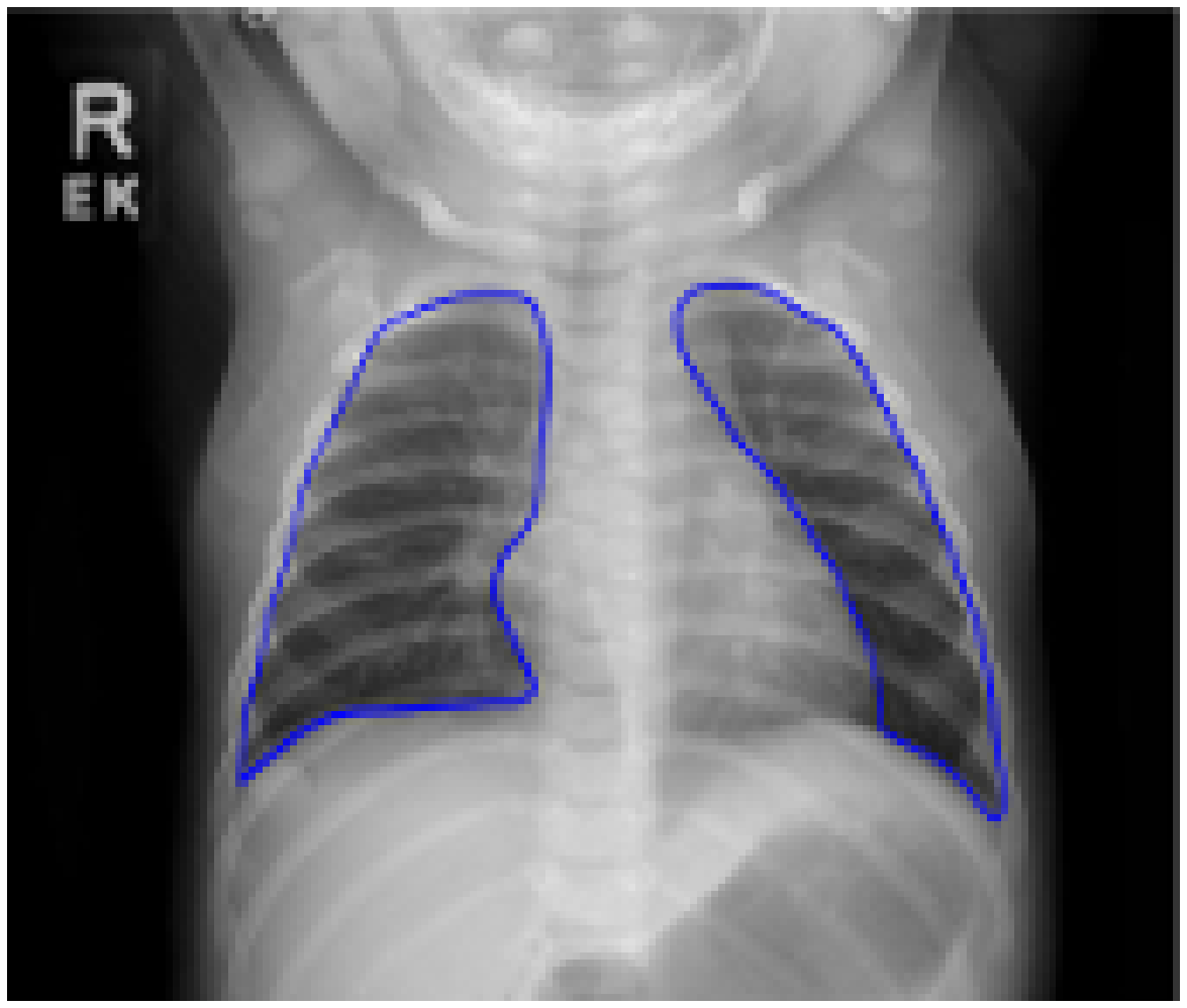}
\caption{}
\label{fig:capacitya}
\end{subfigure}
\end{minipage}
\begin{minipage}{\textwidth}
\begin{subfigure}[c]{0.9\textwidth}
\includegraphics[width=\textwidth]{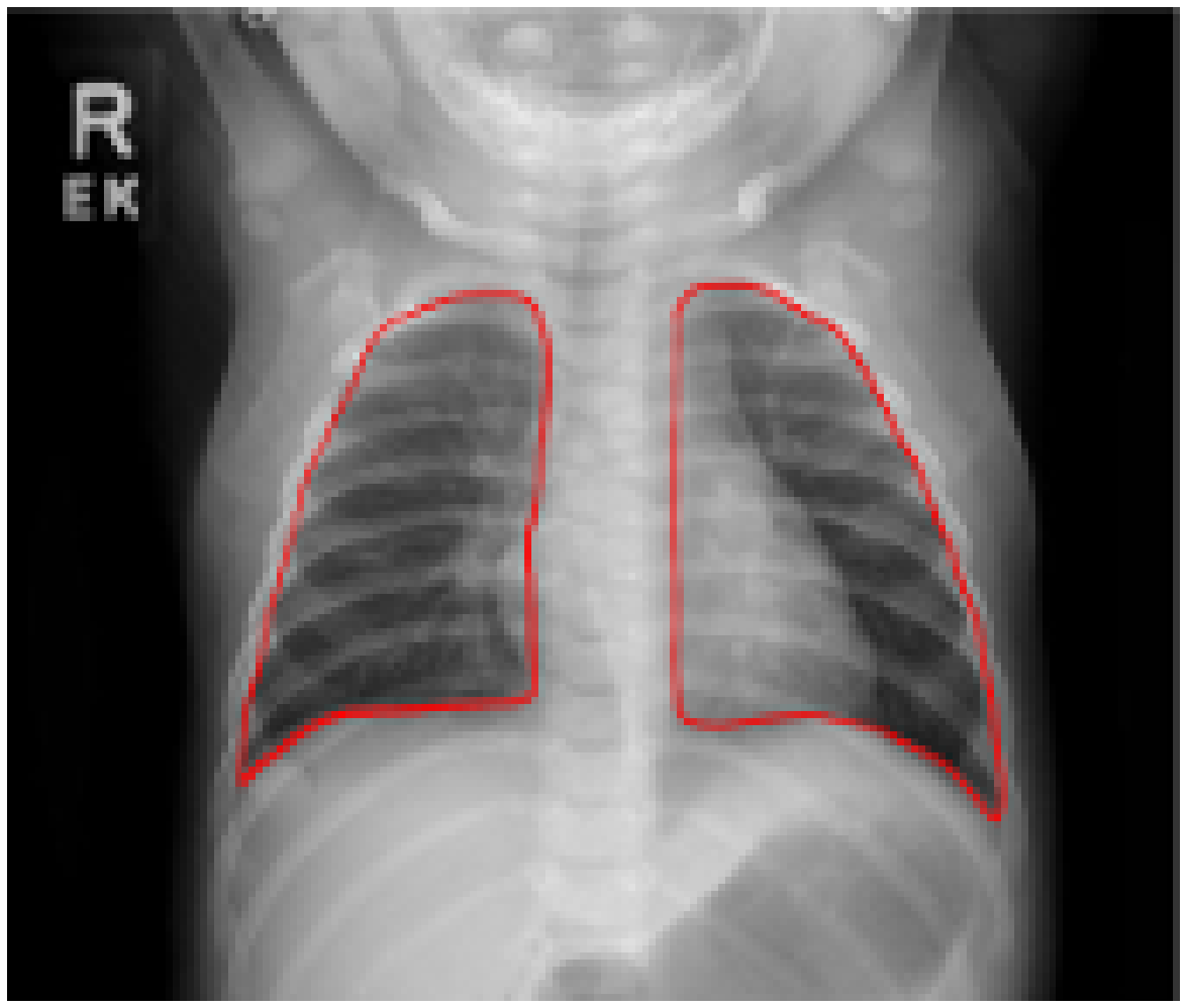}
\caption{}
\label{fig:capacityb}
\end{subfigure}
\end{minipage}
\end{minipage}
\begin{minipage}{.32\textwidth}
\begin{subfigure}[c]{\textwidth}
\includegraphics[width=0.98\textwidth]{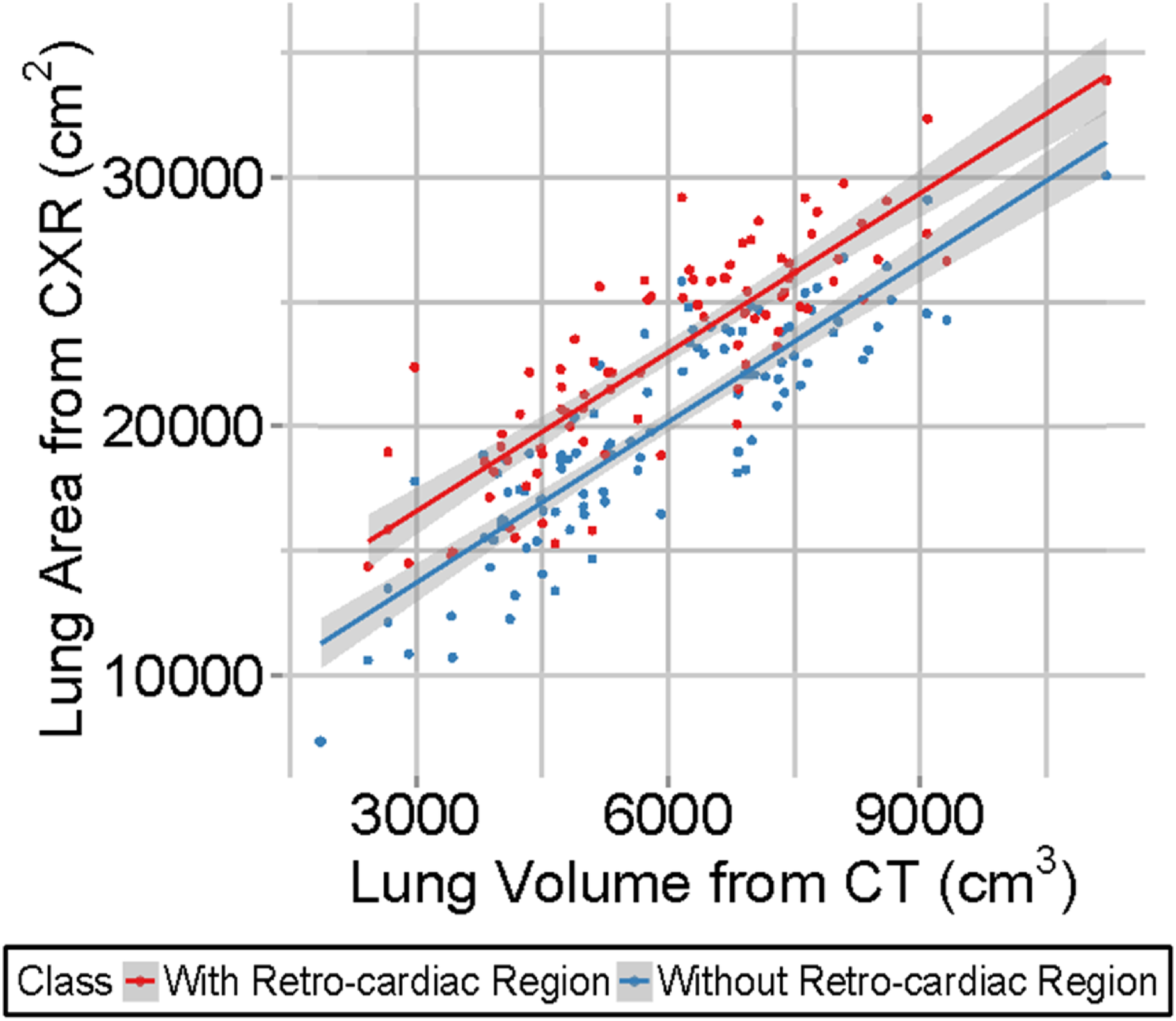}
\caption{}
\label{fig:capacityc}
\end{subfigure}
\end{minipage}
\caption{\footnotesize{A chest radiograph with lung field delineation overlay: (a) without retro-cardiac region, (b) with retro-cardiac region. (c) Correlation between the lung volume estimated from computed tomography (CT) scan with segmented lung field area from CXR (without retro-cardiac region, R=0.80, including retro-cardiac region, R=0.86). Red and blue boundaries indicate the lung field with and without the retro-cardiac region respectively.}}
\label{fig:capacity}
\end{figure}

Traditionally, CAD algorithms designed to segment lung field from CXR ignore the retro-cardiac region, i.e., the lung region occluded by heart (Fig. \ref{fig:capacitya}). The segmentation label without the retro-cardiac region provides only partial unobstructed lung field. Accurate delineation including the occluded retro-cardiac region, is necessary for correct diagnosis in diseases related to the change in lung capacity such as atelectasis (lung collapse), hyaline membrane disease, transient trochnpea, and Meconium aspirat. Fig. \ref{fig:capacityc} presents the correlation between the lung volume estimated from computed tomography (CT) scans and the segmented lung field area from CXR (with and without retro-cardiac region) from 108 individuals. The plot shows a stronger overall correlation between the lung capacity calculated including the retro-cardiac region and the lung volume obtained through CT scans (R=0.80 without retro-cardiac region, R=0.86 including retro-cardiac region; no inspiration/ expiration information was available. R is the correlation coefficient).

The current CXR-based lung segmentation approaches (Table \ref{table:methods}) can be divided into three major categories:\\
\noindent\textbf{Rule-based methods} that use predefined knowledge about the lung field to create a set of rules (e.g. intensity, edge information, etc.) for segmentation. These are usually heuristic approaches therefore subsequent refinement steps are generally needed \cite{brown1998knowledge, duryea1995fully, armato1998automated, li2001improved}.
\begin{table}
\begin{footnotesize}
\caption{\footnotesize{\uppercase{Brief Description of the State-of-the-Art Lung Field Segmentation Techniques.}}\label{table:methods}}
\begin{tabular}{p{2.3cm}p{5.9cm}}
\toprule
\multicolumn{2}{c}{\textbf{Rule-Based Methods}}\\
Brown \emph{et al.} \cite{brown1998knowledge}&Matches the anatomical model of lung to extracted edges from image.\\
Dureya \emph{et al.} \cite{duryea1995fully}&Extracts diaphragm for lung field extraction.\\
Armato \emph{et al.} \cite{armato1998automated}&Uses global and local intensities.\\
Li \emph{et al.} \cite{li2001improved}&Combines edge-based feature classification with iterative contour smoothing.\\
\midrule
\multicolumn{2}{c}{\textbf{Feature Classification-Based Methods}}\\
van Ginneken \emph{et al.} \cite{van2000automatic}&Uses k-NN classifier with Gaussian derivative filters for multiscale pixel classification.\\
McNitt-Gray \emph{et al.} \cite{mcnitt1995feature}&Employes linear discriminator and neural networks with features selected.\\
Dai \emph{et al.} \cite{dai2017scan}&Uses adversarial network that jointly train a segmentation network and a critic network.\\
Wang \emph{et al.} \cite{wang2017segmentation}&Uses fully convolutional network (FCN) to simultaneously segment multiple structures including the lung field within chest radiographs.\\
\midrule
\multicolumn{2}{c}{\textbf{Deformable Shape Model-Based Methods}}\\
Dawoud \emph{et al.} \cite{dawoud2010fusing}& Fuses shape prior with intensity threshold.\\
Annangi \emph{et al.} \cite{annangi2010region}&Integrates lung edge and castrophenic angle into level set information.\\
Sohn \emph{et al.} \cite{sohn2011segmentation}&Uses active contour model \cite{chan2001active} for lung field segmentation.\\
Shi \emph{et al.} \cite{shi2008segmenting}&Combines cohort-specific statistics for constraining the deformable contour.\\
Xu \emph{et al.} \cite{xu2012edge}&Combines edge and region forces for shape model deformation.\\
\midrule
\multicolumn{2}{c}{\textbf{Hybrid Methods}}\\
Shao \emph{et al.} \cite{shao2014hierarchical} & Uses local shape and appearance sparse learning in hierarchical deformation framework.\\
Candemir \emph{et al.} \cite{candemir2014lung}& Uses multiple atlases with non-rigid registration.\\
Ibragimov \emph{et al.} \cite{ibragimov2016accurate}&{\color{black}Employs Haar-like features with random forest classifier to model the appearance of the landmarks and shape-based Gaussian distribution to model the spatial relationship amongst those landmarks.}\\
\bottomrule
\end{tabular}\\
\\
\footnotesize{\textbf{Note:} \emph{None of the methods include the retro-cardiac region as part of the lung field label.}}
\end{footnotesize}
\end{table}

\noindent\textbf{Feature classification-based methods} that formulate segmentation as a classification problem by learning the probability of every pixel (or region) belonging the lung field. The probability is calculated using a set of features extracted around the pixel being classified \cite{van2000automatic, mcnitt1995feature}. Recently, \cite{dai2017scan} used adversarial architecture for the lung field segmentation. Adversarial networks are generally harder to train, i.e., large datasets and exhaustive parameter optimization is needed. Furthermore, as demonstrated later in Section \ref{sec:results} (Experimental Results) ignorance of the object shape specificity results in the suboptimal performance even by the most sophisticated feature classification-based methods.

\noindent\textbf{Deformable shape model-based methods} that use curves and surfaces defining the lung field that can be moved to the true boundary under the influence of \emph{internal forces} from lung shape and \emph{external forces} from lung appearance \cite{dawoud2010fusing, annangi2010region, sohn2011segmentation, shi2008segmenting, xu2012edge}.

In addition, hybrid methods such as \cite{shao2014hierarchical} and \cite{candemir2014lung} cross over multiple categories. Amongst these approaches, deformable statistical shape models (SSMs) have demonstrated superior performance due to their ability to seamlessly integrate low level localized appearance features and high level global features. These models learn patterns of shape deformation from the training data of annotated images. A learned model is subsequently deformed to fit the object of interest within the test image by estimating its shape deformation patterns through an appearance-guided iterative optimization procedure. SSMs remained the workhorse for various medical image analysis applications including the lung segmentation; however, the iterative optimization is generally found to be not robust to initialization, complex background, weak edges, and contrast information. Henceforth, accurate initialization of shape models \cite{cosio2008automatic} and various refinements \cite{zhang2012towards} remain topics of active research. In addition, conventional SSMs \cite{cootes1995active}, assume a unimodal Gaussian distribution of training shapes; however, in practice, the assumption of both unimodality as well as Gaussianity may be inaccurate when the training data consist of shapes with large variation obtained from multiple cohorts, e.g., from adult and pediatric subjects (see Fig. \ref{fig:adultPedsPCA}, \ref{fig:adultPedsAspectRatio}). 

Contrary to SSM methods, representation learning techniques have demonstrated great potential in handling a wide range of variation including non-Gaussian and multi-modal Gaussian distributed data \cite{zheng20081668, ghesu2015marginal, 7422082}. These techniques have also found to be robust to intensity variation and minima optimization. However, the cost of performing hypothesis testing at the atomic (pixel/voxel) level prohibits their use for large object segmentation. Furthermore, since final segmentation label using these methods are generally obtained as a concatenation of independent atomic-level hypotheses, object shape specificity cannot be guaranteed. {Shape modeling through representation learning has not garnered much attention in the past, primarily because of two reasons. First, the effective representation of a segmentation (detection+delineation) task as a learning problem is not trivial. Second, hand-crafting representation features for deformable objects is not straightforward and relies heavily on the human ingenuity} \cite{zheng20081668, ghesu2015marginal}. 

Recently, representation learning through deep learning (DL) has shown great promise in expanding the scope of learning algorithms to automated feature extraction. Specific to medical imaging, DL frameworks are extensively being used in various organ detection \cite{shin2013stacked}, classification \cite{7422082}, and segmentation \cite{ronneberger2015u} tasks. In this paper, we extend the applicability of DL to parametrized shape learning and demonstrate it via an efficient generic solution to lung field segmentation. {The main contributions of our work are:}
\begin{itemize}
\item {A generic lung field segmentation framework from CXR, accommodating both adult and pediatric cohorts.}
\item {Segmentation of the lung field including the occluded retro-cardiac region for reliable estimation of capacity and inter/intra subject comparisons.}
\item A DL-based mechanism for the automated detection of object of interest with large shape variation from images acquired under diverse acquisition protocols. This detection mechanism, dubbed \emph{ensemble space learning} (ESL), also addresses the issue of error propagation to subsequent marginal spaces within the current state-of-the-art detection methods: marginal space learning (MSL) \cite{zheng2008four, ghesu2016marginal}. 
\item A hybrid principal component analysis (PCA)-DL based approach for including shape prior information for deformable object segmentation. This module which we call \emph{marginal shape deep learning} (MaShDL) transforms the iterative approach of the conventional SSM-based segmentation methods to a recursive marginal refinement approach. Specifically, the method begins by learning the mode of shape deformation in the eigenspace of the largest variation and then marginally increases the dimensionality of eigenspaces by recursively including the next largest modes. As demonstrated later in the paper, the transformation allows the SSM to be posed as an efficient parameter estimation problem solvable through representation learning.
\end{itemize}
The proposed framework is evaluated using a comprehensive CXR datasets to demonstrate its potential for generic applicability.

\section{Datasets and Reference Standards}
\label{sec:data}
Our experiments are conducted on both publicly available and in-house acquired datasets using a wide range of devices, age groups, and pulmonary pathologies. 247 publicly available radiographs from Japanese Society of Radiological Technology (JSRT; \url{http://www.jsrt.or.jp}) dataset and 108 from the Belarus Tuberculosis Portal (BTP; \url{http://tuberculosis.by}) were used. For data acquired in-house, after approval from the Internal Review Board, 313 posterior-anterior CXRs were collected at Children's National Health System (CNHS). The subjects in the JSRT dataset have ages between 16 to 89 year (\begin{small}$58.21\pm 14.02$\end{small} year). The dataset is a standard digital CXR database with and without chest lung nodules created by the Japanese Society of Radiological Technology.
The radiographs had dimensions of \begin{small}$2048\times 2048$\end{small} pixels, spatial resolution of \begin{small}$0.17\times 0.17$\end{small} mm/pixels, and digital resolution of 12 bits. BTP images, from patients between 18 to 86 year (\begin{small}$45.60\pm 16.98$\end{small} year), had dimensions of \begin{small}$2248\times 2724$\end{small} pixels, spatial resolution of \begin{small}$0.16\times 0.16$\end{small} mm/pixel and the digital resolution of 12 bits. The dataset consists of CXRs obtained from patients diagnosed with or suspected of multi-drug-resistant tuberculosis (MDR-TB). The CXR findings of these patients include consolidation, cavitary lesions, nodules, pleural effusion, pneumothorax, and fibrotic scars. For CNHS data, patients having ages between 3 months to 18 year (\begin{small}$4.75\pm 5.30$\end{small} year) with viral chest infections were scanned. The dataset consists of radiographs collected from individuals having or suspected of having either Human metapneumovirus (hMPV) or rhinovirus. The radiological symptoms to these viruses include acute respiratory infections, chronic lung conditions, chest wall deformities, cardiovascular anomalies. The radiographs have dimensions within the range \begin{small}$\left(660-4240\right)\times\left(987-4240\right)$\end{small} pixels with spatial resolution ranges between \begin{small}$0.1\times 0.1\text{mm/pixel}$\end{small} to \begin{small}$0.14\times 0.14\text{mm/pixel}$\end{small} and a digital resolution of 12 bits. For CNHS data, patients having ages between 3 months to 18 year (\begin{small}$4.75\pm 5.30$\end{small} year) with viral chest infections were scanned. The dataset consists of radiographs collected from individuals having or suspected of having either Human metapneumovirus (hMPV) or rhinovirus. The radiological symptoms to these viruses include acute respiratory infections, chronic lung conditions, chest wall deformities, cardiovascular anomalies. For consistency of training data, all scans from the three datasets were resized to \begin{small}$2048\times 2048$\end{small} pixels using B-spline interpolation.

The ground truth labels both including and excluding the retro-cardiac region were prepared by two fellows using the ITK-SNAP interactive software (~\href{http://www.itksnap.org}) under the supervision of two expert pulmonologists. {For ground truth labels including the retro-cardiac region, an overall inter-observer agreement of $0.95\pm 0.03$ was observed; specifically, $0.94\pm 0.02$ for CNHS data and $0.96\pm 0.03$ for the JSRT and BTP data was estimated. Ground truth labels excluding the retro-cardiac region were prepared for comparative purposes with the state-of-the-art methods.} To construct the statistical shape model, 144 boundary points (72 per left/right lung) with anatomical correspondences are annotated. Specifically, six manually annotated primary landmarks were initially obtained for each lung based on their distinctive anatomical appearance and ability to roughly define the shape of lung. Subsequently, equidistant secondary landmarks were estimated along the lung contour using interpolation between the primary landmarks. In order to make sure that no loss in the segmentation label accuracy has occurred due to the interpolation, the accuracy of the proposed interpolation method was evaluated using the Dice coefficient score (DCS) between the manual ground truth and the landmark-based interpolated contour. A mean DCS of $0.9942\pm0.0013$ was obtained for our dataset. Further details on our manual landmarking approach can be found in \cite{okada2015severity}.

\section{Methods}
\subsection{Overview}
\begin{figure}
\centering
\includegraphics[width=0.45\textwidth]{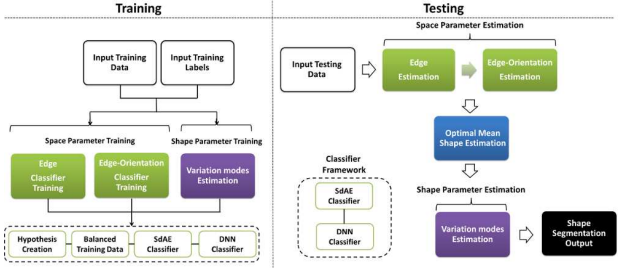}
\caption{\footnotesize{The overview flow diagram of the proposed method for generic space and shape learning.}}
\label{fig:overviewMain}
\end{figure}

Fig. \ref{fig:overviewMain} shows the flow diagram summarizing the proposed framework. The segmentation of a deformable object (lung field) is performed by learning space (localization) and shape parameters using two separate DL architectures. As demonstrated later in the manuscript, the presented DL-based approach for shape parameters learning is theoretically equivalent to the one adopted by conventional SSM techniques: estimating the shape parameters of the object of interest under constraints on shape model and appearance. However, unlike the iterative convergence approaches of conventional SSMs that optimize the entire shape parameter space simultaneously, the proposed method transforms the parameter space into linearly independent subspaces and employs a battery of DL classifiers to learn the shape parameters individually. This marginal learning of independent parameter subspaces makes our approach both computationally tractable as well as significantly more accurate compared to the state-of-the-art SSM approaches. Herein, we introduce a generic method for space and shape parameters learning of deformable objects, which we apply later to the lung field segmentation from CXR.

\subsection{Parametrized Shape Representation and Learning}
\label{sec:parametrized}
Among the approaches for deformable shape representation presented in the literature \cite{davies2008statistical}, PCA-based SSM \cite{cootes1995active} has been found to be most successful due to their simplicity, performance, and compact representation. These models have been widely used to deform an initial estimate of shape (mostly the mean shape of the object of interest obtained using training data) under the guidance of appearance-based image evidence (\emph{external forces}) and shape priors (\emph{internal forces}). SSM uses an explicit point-based representation in which each shape is described by $M$ points (or landmarks) distributed across the contour. Given a set of \begin{small}$N$\end{small} aligned shapes $\{\mathbf{x}_n\}_{n=1}^N$ in 2D, the SSM is defined using a mean shape $\overline{\mathbf{x}}\in\mathbb{R}^{2M\times 1}$, a set of $K$ eigenvectors $\{\mathbf{p}_k\}_{k=1}^K$, and a set of corresponding eigenvalues, $\{\lambda_k\}_{k=1}^K$, obtained by applying PCA to the aligned shapes. The magnitude is proportional to the shape variance explained by the particular eigenvector. $K$ is generally chosen to be the smallest number of modes such that their cumulative variance explains a sufficiently large proportion (normally \begin{small}$95\%-98\%$\end{small}) of the total variance explained by all $M$ eigenvectors \begin{small}(usually $K\ll M$)\end{small}. Subsequently, any shape \begin{small}$\mathbf{X}\in\mathbb{R}^{2M\times 1}$\end{small} in the non-aligned image space can be approximated using the anisotropic similarity transform parameters (presented below), the aligned mean shape \begin{small}$\overline{\mathbf{x}}$\end{small}, and the weighted sum of \begin{small}$K$\end{small} largest modes (eigenvectors).

\begin{small}
\begin{equation}
\mathbf{X} \approx A_\text{space} \left(\overline{\mathbf{x}} + {{\mathbf{p}}{\mathbf{b}}}\right),
\label{eq:asm}
\end{equation} 
\end{small}

\noindent where \begin{small}$A_\text{Space}=\left[ {\begin{array}{*{20}{c}}
{\mathbf{S}\theta}&\mathbf{T}\\
{{0^T}}&1
\end{array}} \right]$\end{small} is an invertible matrix called the anisotropic similarity transform matrix. The matrix transforms the mean shape from the aligned shape space to a non-aligned image space, using specifically, position: \begin{small}$\mathbf{T}=\{\mathbf{T}_x, \mathbf{T}_y\}$\end{small}, orientation: $\theta$, and anisotropic scale: \begin{small}$\mathbf{S}=\{\mathbf{S}_x, \mathbf{S}_y\}$\end{small}. $\mathbf{b}\in\mathbb{R}^{K\times 1}$ is the shape weight matrix. 

Given $A_\text{space}$ and \begin{small}$\overline{\mathbf{x}}$\end{small}, the weight parametrization of new target shape can be obtained using (\ref{eq:asm}) as \begin{small}$\mathbf{b}=\mathbf{p}^T\left(A_\text{space}^{-1}\mathbf{X}-\overline{\mathbf{x}}\right)$\end{small}. The legitimacy of the estimated shape is generally guaranteed by imposing individual constraints on each weight. \cite{cootes1995active} demonstrated that the suitable constraints on the weights are typically of the order of \begin{small}$- 3\sqrt {{\lambda _k}}\le {\mathbf{b}_k} \le+3\sqrt {{\lambda _k}}$\end{small}.

After initializing the similarity parameters described by $A_\text{space}$, SSM iteratively adjusts the deformable shape until convergence, causing the points of $\mathbf{X}$ to move under the influence of object model and image evidence. The weight vector ${\mathbf{b}}$ after $t$ iterations is

\begin{small}
\begin{equation}\mathbf{b}^{(t)} \leftarrow \mathbf{b}^{(t-1)} + d\mathbf{b}^{(t)}, \label{eq:modeweights}\end{equation}\end{small}

\noindent where $d\mathbf{b}^{(t)}=\mathbf{p}^Td\mathbf{x}$ is the change in the model parameters at the iteration $t$. Using eq. (\ref{eq:asm}) and (\ref{eq:modeweights}), the shape parameter vector $\Omega$ for any deformable shape in 2D can be written as 

\begin{small}\begin{eqnarray}\Omega=\left\{ {\begin{array}{*{20}{c}}
\underbrace{ \mathbf{T} \in {\mathbb{R}^2},\theta \in {\mathbb{R}},\mathbf{S} \in {\mathbb{R}^2}}_\text{Space Parameters},
\underbrace{\overline{\mathbf{x}}\in\mathbb{R}^{2M\times 1}}_\text{Mean Shape}, \underbrace{\mathbf{b} \in {\mathbb{R}^{K\times 1}}}_\text{Shape Parameters}
\end{array}} \right\}
\label{eq:parameters}
\end{eqnarray}
\end{small}

Given a shape parameter vector $\Omega$ and a set of 2D training images $\{f_n\}_{n=1}^N$, each having dimensions \begin{small}$(Nx, Ny)$\end{small}: \begin{small}$f(nx,ny)\in\mathbb{R}: 0\le nx\le Nx-1,~0\le ny\le Ny-1$\end{small} ({where $f(nx,ny)$ denotes intensity at location $(nx, ny)$), a representation classifier can be learned that can estimate the correct parameter vector, $\Omega$, by maximizing the following posterior probability over a valid parameter space}:

\begin{small}
\begin{equation}\widehat{{\Omega}} = \{\underbrace{\widehat{\mathbf{T}}, \widehat{\theta}, \widehat{\mathbf{S}}}_{\Omega_\text{Space}}, \underbrace{{\overline{\mathbf{x}}}}_{\Omega_\text{Mean Shape}}, \underbrace{\widehat{\mathbf{b}}}_{\Omega_\text{Shape}} \}=\mathop {{\rm{arg max}}}\limits_{\Omega} p\left({{\Omega}|f} \right).\label{eq:localization}\end{equation}
\end{small}
However, due to the large number of testing hypotheses in eq. (\ref{eq:localization}), learning a classifier with efficiency comparable to the traditional SSM-based iterative segmentation methods is challenging and requires a large amount of training data. A few attempts have been made in the past for efficient parameter learning by partitioning the parameter space into linearly or marginally independent subspaces. For instance, \cite{zheng20081668} proposed an efficient method, MSL, for object detection by training classifiers to learn $\Omega_\text{Space}$. Since its introduction, MSL has been successfully applied in various medical imaging applications such as segmentation of heart \cite{zheng20081668}, left ventricle detection \cite{zheng2009robust}, mid-sagittal plane detection \cite{schwing2013method}, and standard echocardiographical plane detection \cite{lu2008autompr}. MSL learns classifiers in marginally independent parameter subspaces. Their work suggested that the dimensionality of effective parameter space can be significantly contracted by separating conditionally independent parameters into semigroups (translations, scales, and orientations). A semigroup is an algebraic structure consisting of a set with an associative binary operation. According to MSL, the object detection approach can be expressed as the maximization of posterior probability of semigroup $\Omega_\text{Space}$,
\begin{equation}p(\Omega_\text{Space}|f) = \mathop {{\rm{arg max}}}\limits_{\Omega_\text{Space}}p\left( {\mathbf{T}|f} \right)p\left( {\theta |\mathbf{T},f} \right)p\left( {\mathbf{S}|\mathbf{T},\theta ,f} \right).
\label{eq:msl}
\end{equation}

Extending the concept, we propose that the posterior probability of the semigroup $\Omega$ can be similarly approximated as the maximization of the marginal probabilities of its semisubgroups: $\Omega_\text{Space}$ and $\Omega_\text{Shape}$,

\begin{small}
\vspace{-0.1in}
\begin{eqnarray}
\mathop {{\rm{arg max}}}\limits_{\Omega}p\left({{\Omega}|f} \right)&=\mathop {{\rm{arg max}}}\limits_{\Omega_\text{Shape}} p\left({{\Omega}_\text{Shape}|{\Omega}_\text{Space}, f} \right)\times \nonumber\\&\mathop {{\rm{arg max}}}\limits_{\Omega_\text{Space}}p\left({{\Omega}_\text{Space}|f} \right),
\label{eq:marginal01}
\end{eqnarray}
\end{small}
However, in contrast to eq. (\ref{eq:msl}) and the MSL framework proposed in \cite{zheng20081668} that does not impose any commutativity constraints, our proposition in eq. (\ref{eq:marginal01}) is subject to an assertion that the parameter vector $\Omega$ can be estimated marginally only as a nowhere-commutative semigroup: \begin{small}$\{\Omega_\text{Space}, \Omega_\text{Shape}|~\Omega_\text{Space}.\Omega_\text{Shape}\ne\Omega_\text{Shape}.\Omega_\text{Space}\}$\end{small}. A nowhere commutative semigroup is any semigroup $S$, such that for all $a$ and $b$ $\in S$ if $ab=ba$ then $a=b$. The nowhere-commutativity is enforced since, as discussed before, within the context of SSM-based methods (eq. (\ref{eq:asm})), image-aligned mean shape ($\overline{\mathbf{X}}=A_\text{space}\overline{\mathbf{x}}$) serves as the initialization for the shape deformation. During the iterative process of (\ref{eq:modeweights}), this initial estimate is continuously refined till convergence.

The marginal parameter space simplification introduced in eq. (\ref{eq:marginal01}) is mainly intended to improve the computational cost of a classifier-based approach to parameter estimation. However, despite this simplification, the proposed classifier-based framework meet or exceed the iterative refinement-based SSM alternatives in terms of segmentation accuracy as demonstrated later in Section \ref{sec:mashdlasm}. The semisubgroups $\Omega_\text{Space}$ and $\Omega_\text{Shape}$ can be further partitioned till the trivial semisubgroup level, i.e., a semigroup with one element only, 

\begin{small}
\begin{eqnarray}\widehat{{\Omega}}_\text{Space} \mathop=\limits^{c} \{\widehat{\mathbf{T}}, \widehat{\theta}, \widehat{\mathbf{S}}\}=&\mathop {{\rm{arg max}}}\limits_{\Omega_\text{Space}} p\left({{\Omega_\text{Space}}|f} \right)\nonumber\\
=&\underbrace{\underbrace{p\left({\mathbf{T}|f} \right)}_{\text{Position}}\times p\left({\mathbf{S}|\mathbf{T},f} \right)}_{\text{Position-Scale}}\times p\left({\theta|\mathbf{T},\mathbf{S},f} \right),
\label{eq:rigid}\end{eqnarray} 
\end{small}
\noindent and $\widehat{{\Omega}}_\text{Shape}$
\begin{small}
\begin{eqnarray}\widehat{{\Omega}}_\text{Shape}\mathop=\limits^{nwc}\{\widehat{\mathbf{b}}_k:\forall k=1,\dots,K\}=\mathop {{\rm{arg max}}}\limits_{\Omega_\text{Shape}} p\left({{\Omega_\text{Shape}}|{\Omega_\text{Space}}, f} \right)\nonumber\\
=\underbrace{\underbrace{p\left( {\mathbf{b}_1|\Omega_\text{Space},f} \right)}_{\text{Largest mode of shape deformation}}\times \dots \times p\left( {\mathbf{b}_K|\mathbf{b}_{K-1},\dots,\mathbf{b}_1,\Omega_\text{Space},f} \right)}_{\text{$K$ largest modes of shape deformation}},
\label{eq:nonrigid}
\end{eqnarray}
\end{small}
\noindent where $c$ and $nwc$ denote commutative and nowhere-commutative semisubgroups respectively. Also note that $p\left( {\mathbf{b}_0|\Omega_\text{Space},f} \right)=\overline{\mathbf{x}}$. Eq. (\ref{eq:rigid}) and (\ref{eq:nonrigid}) suggest that by splitting the semigroups $\Omega$ to commutative and non-commutative non-trivial semisubgroups, a $5+K$ dimensional learning space can be approximated to a concatenation of $5+K$ one dimensional subspaces; therefore, reducing the computational complexity of the manifold \cite{zheng2008four, ghesu2016marginal}. Individual classifiers can be trained subsequently for independent subspaces, thus simplifying training and reducing the amount of data needed to train the classifier.

\subsection{Deep Learning Network for Space and Shape Parameter Estimation}
\label{sec:sdae}
{Our proposed DL framework for learning the parameters $\Omega$ consists of two main layers: an unsupervised stacked denoising autoencoder (SdAE) layer for pre-training to initialize the weights of feed forward deep neural network (DNN) and a supervised DNN layer for fine-tuning. Unsupervised pre-training to initialize the weights of DNN has demonstrated to have better convergence properties especially if the labeled training data is not very large} \cite{erhan2010does}. 

An autoencoder (AE) consists of two components: the encoder and the decoder. The AE, in our framework, takes a vectorized $q\times q$ image patch(es) as input ${f}_{q^2\times 1}\in[0,1]^{L_1}$ and maps it to a hidden representation $\mathbf{h}\in[0,1]^{L_2}$ through a deterministic mapping, $\mathbf{{h}}=\sigma \left( {\mathbfcal{W}_\textbf{h} {{f}}+{\beta_1}} \right)$, where $\mathbf{{h}}$ is called the activation vector, $\sigma$ is the logistic sigmoid function $\sigma \left( a \right) = {\left( {1 + \exp \left( { - a} \right)} \right)^{ - 1}}$, $\mathbfcal{W}_\textbf{h}\in\mathbb{R}^{L_2\times L_1}$ is the mapping matrix, and ${\beta_1}\in\mathbb{R}^{L_1}$ is the bias vector. The decoder maps back to the same shape as the observed data using the reverse mapping. The denoising autoencoder (dAE) is a stochastic version of the AE. Specifically, to force the hidden layers to discover more robust features, the dAE is trained to reconstruct the input from its corrupted version. Finally, the SdAE \cite{larochelle2007empirical} is a DNN consisting of multiple layered dAEs. 

Once the layers are pre-trained using SdAE, the weights and biases of the encoder layer are used to initialize the feed forward DNN. This network architecture is subsequently used for learning space and shape parameters in our DL framework. For greater details on the training of DNNs and SdAE, readers are encouraged to review \cite{vincent2008extracting}. Specific details of the network configuration pertaining to learning space and shape parameters are presented in Sections \ref{sec:esl} and \ref{sec:mashdl} respectively.
\subsection{Space Parameters Estimation}
\label{sec:esl}
\begin{figure}
\begin{subfigure}[c]{0.22\textwidth}
\centering
\includegraphics[width=\textwidth, height=3.75cm]{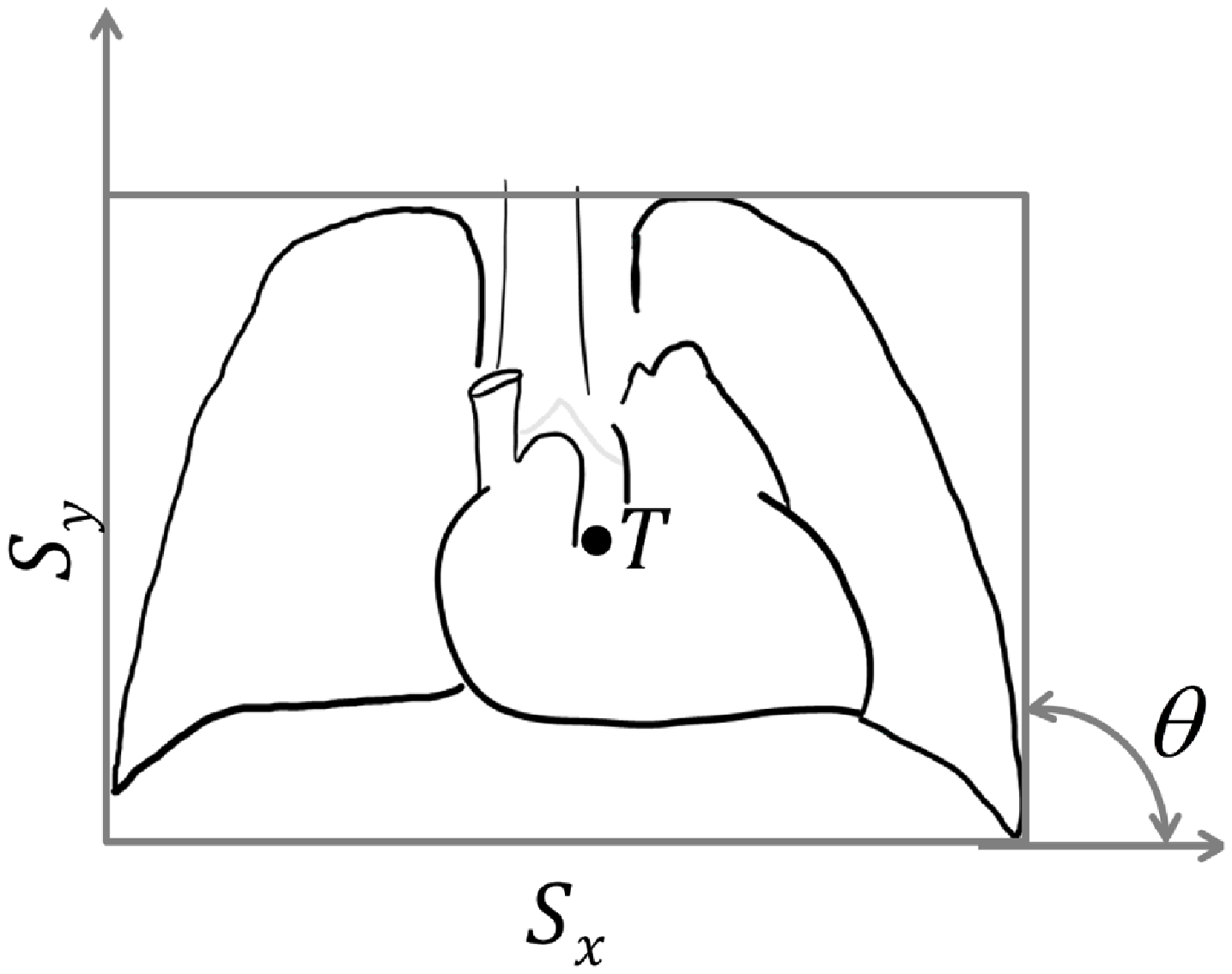}
\caption{}
\label{fig:msl}
\end{subfigure}
\begin{subfigure}[c]{0.22\textwidth}
\centering
\includegraphics[width=\textwidth, height=3.75cm]{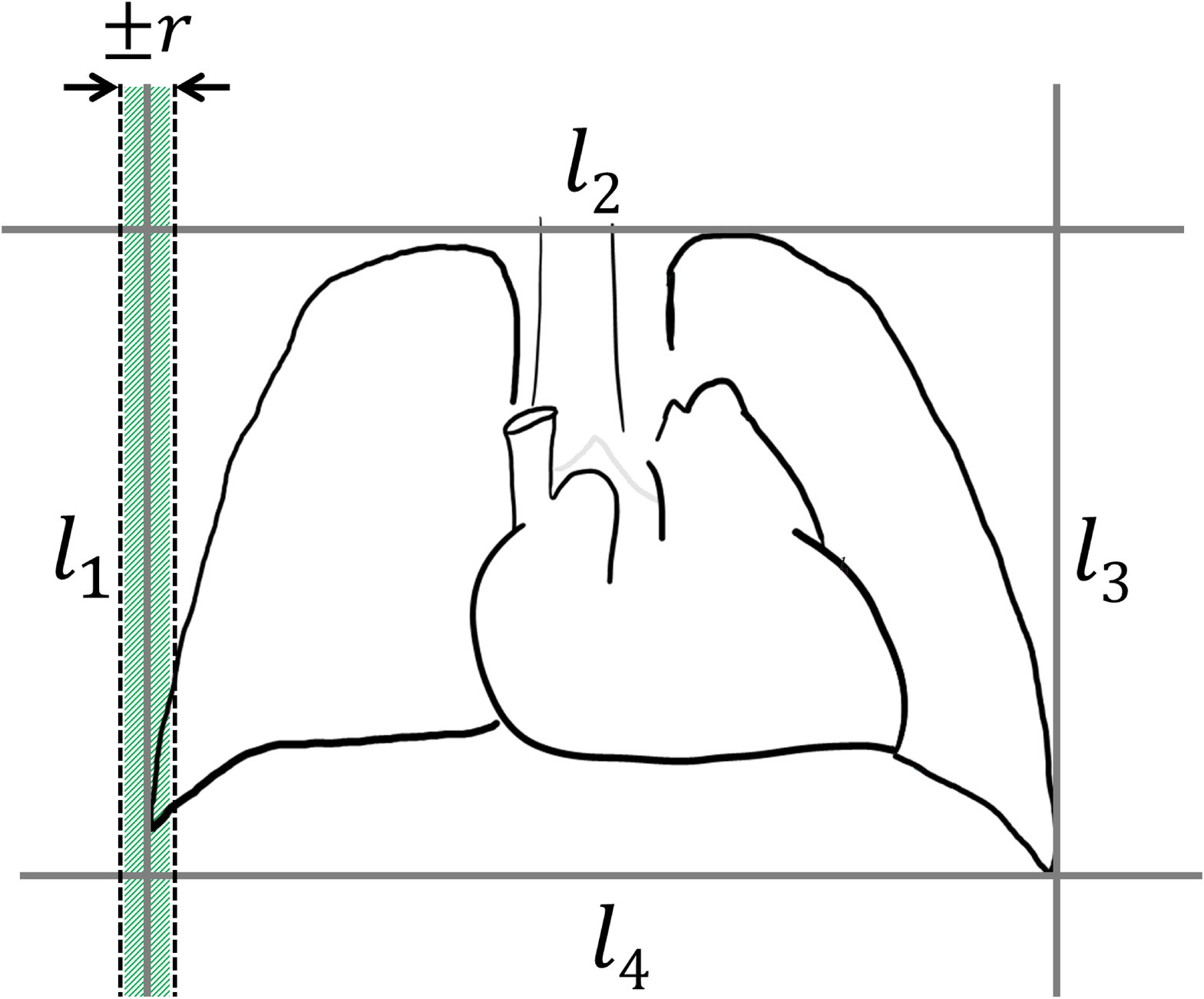}
\caption{}
\label{fig:esl}
\end{subfigure}
\caption{\footnotesize{Illustration of the differences in approach between the marginal space learning (MSL) and the ensemble space learning (ESL) to estimate $\Omega_\text{Space}$. (a) MSL, (b) ESL. To estimate $\Omega_\text{Space}$, MSL uses minimum area bounding box while ESL employs linearly independent bounding lines. {The dashed green patch of size (}$2r+1$, $Ny${) shows a positive hypothesis satisfying eq.} (\ref{eq:transPositive}) {for line} $l_1$.}}
\label{fig:eslmsl}
\end{figure}

MSL and MSDL, the current state-of-the-art learning-based techniques for space parameters estimation ($\widehat{\Omega}_\text{Space}$), has been found to be very successful in various medical imaging applications \cite{zheng2014marginal, ghesu2015marginal}. Both approaches solve the same classification problem using two different classification techniques. MSL uses the probabilistic boosting tree classifier while MSDL adopts the deep neural network for the parameter estimation. Both MSL and MSDL are initialized using a bounding box of arbitrary parameters (Fig. \ref{fig:msl}). Later these parameters are marginally refined (translation followed by orientation followed by scale). The marginal refinement transforms the arbitrary bounding box into a minimum area bounding box enclosing the object of interest. The sequential parameter learning within MSL, however, results in the propagation of estimation error to successive stages. Specifically, the error in the translation estimation propagates to orientation and scale estimations. Consequently, the cumulative estimation error at a given stage is lower-bounded by the cumulative error at the previous stages,

\begin{small}
\vspace{-0.1in}
\begin{eqnarray}
\inf \left(\Omega_\text{Translation}-\mathop {{\rm{arg max}}}\limits_{\mathbf{T}} p\left({{\mathbf{T}}|f} \right)\right)\rightarrow 0\nonumber,\\
\inf \left(\Omega_\text{Translation-Orientation}-\mathop {{\rm{arg max}}}\limits_{\mathbf{T,\theta}} p\left({{\mathbf{T}},\theta|f} \right)\right)\rightarrow\mathbf{T}-\widehat{\mathbf{T}}\nonumber,\\
\inf \left(\Omega_\text{Space}-\mathop {{\rm{arg max}}}\limits_{\mathbf{T,\theta,S}} p\left({{\mathbf{T,\theta,S}}|f} \right)\right)\rightarrow\theta-\widehat{\theta}.
\label{eq:rigid01}
\end{eqnarray}
\end{small}

From eq. (\ref{eq:rigid01}), the domain normalized error \begin{small}$\mathbf{S}-\widehat{\mathbf{S}}\ge \theta-\widehat{\theta}\ge\mathbf{T}-\widehat{\mathbf{T}}$\end{small} (further explanation on the propagation of error is provided in Section \ref{sec:mslvesl}). Moreover, since MSL and MSDL are based on using a minimum area bounding box, deciding the optimal initialization values of similarity transform parameters ($\Omega_\text{Space}$) for the bounding box is generally not trivial, especially in data with large variation. To address these challenges, we propose ESL that learns $\Omega_\text{Space}$ by transforming it from being a marginally independent semigroup of parameters (as described in MSL, eq. (\ref{eq:rigid01})), to a linearly independent semigroup of surrogate parameters. Specifically, instead of estimating $\Omega_\text{Space}$ using the minimum area bounding box, ESL estimates them as a function of four linearly independent vertices of two sets of parallel lines bounding the object of interest. Fig. \ref{fig:eslmsl} graphically illustrates the methodological differences between MSL and ESL for the specific application of lung field segmentation. Given a pair of parallel bounding lines $l_{\{1,2\}}\in\mathbb{R}^2$, and a second pair of lines $l_{\{3,4\}}\in\mathbb{R}^2$ perpendicular to $l_{\{1,2\}}$, the four intersecting vertices provide the estimation of translation ($\mathbf{T}$) and the scale ($\mathbf{S}$) of the minimum area bounding box enclosing the object of interest (lung field). The box of estimated translation and scale is subsequently used to estimate the orientation ($\theta$). Unlike the MSL, no assumption on the initial values of parameters is needed in the ESL. Moreover, since the parameters of ESL are linearly independent (i.e., \begin{small}$p\left({{l_i}|{l_j},f} \right)=p\left({{l_i}|f} \right), \forall i\neq j$\end{small}; in MSL (\begin{small}$p\left({\mathbf{S}|\mathbf{T},f} \right)\neq p\left({\mathbf{S}|f} \right)$\end{small})); therefore, \begin{small}$p\left({{\Omega}_\text{Space}|f} \right)=\mathcal{L}\{\prod_{i=1}^4 p(l_i|f)\}\times p\left({\mathbf{\theta}|\mathbf{T},\mathbf{S},f} \right)$\end{small}, where $\mathcal{L}\{\}$ denotes the sequence of geometrical operations to extract $\mathbf{S}$ and $\mathbf{T}$ from the four estimated vertices. Similar to eq. (\ref{eq:rigid01}), the lower-bounds on cumulative estimation error for the space parameters using the ESL is, 

\begin{small}
\vspace{-0.1in}
\begin{eqnarray}
\inf \left(\Omega_\text{Translation}-\mathop {{\rm{arg max}}}\limits_{\mathbf{T}} p\left({{\mathbf{T}}|f} \right)\right)\rightarrow 0\nonumber,\\
\inf \left(\Omega_\text{Translation-Scale}-\mathop {{\rm{arg max}}}\limits_{\mathbf{T,S}} p\left({{\mathbf{T}},\mathbf{S}|f} \right)\right)\rightarrow 0\nonumber,\\
\inf \left(\Omega_\text{Space}-\mathop {{\rm{arg max}}}\limits_{\mathbf{T,\theta,S}} p\left({{\mathbf{T,\theta,S}}|f} \right)\right)\rightarrow l_i-\widehat{l}_i.
\label{eq:eslerror}
\end{eqnarray}
\end{small}
Since the orientation is estimated independently and PA CXRs are acquired under a position protocol (upright), pairs $l_{\{1,2\}}$ and $l_{\{3,4\}}$ can be assumed to be parallel to the horizontal axis and the vertical axes of the image respectively for simplicity (Fig. \ref{fig:esl}). Therefore, for the pairs of lines bounding the object of interest parallel to the horizontal ($i\in\{1,2\}$) and vertical ($i\in\{3,4\}$) axes, the bounding line estimation problem is reduced to estimating two pairs of x-intercepts (lines bounding the object vertically: $l_1$, $l_3$) and y-intercepts (lines bounding horizontally: $l_2$, $l_4$).

\subsubsection{Bounding Lines Estimation}\label{sec:edge}
\noindent\textbf{Training:} Four separate DL classifiers are trained for the four bounding lines. To provide contextual information to the classifier, an image patch of size (\begin{small}$2r+1$, $Ny$\end{small}) or (\begin{small}$Nx$, $2r+1$\end{small}) are extracted around each line: \begin{small}$\{l_i\}_{i=1}^4\pm r$\end{small} (see Fig. \ref{fig:esl}). A positive hypothesis for a line $l_i$ is formulated to find the horizontally (or vertically) oriented bounding box centered at position \begin{small}$(\{l_i\}_{i=\{1,3\}}, \frac{Ny}{2})$\end{small} (or \begin{small}$(\frac{Nx}{2}, \{l_i\}_{i=\{2,4\}})$\end{small}) respectively,
\begin{equation}
|l_i-\widehat{l_i}|\le 1 \text{ pixels},
\label{eq:transPositive}
\end{equation}
where $l_i$ and $\widehat{l_i}$ denote the ground truth and the hypothesized position of the line $i$ respectively. Similarly, a negative sample satisfies: 
\begin{equation}
|l_i-\widehat{l_i}|\ge 5 \text{ pixels},
\label{eq:transNegative}
\end{equation} 
The separation in the positive and negative hypotheses is intended to provide a clean split between the training hypotheses. 

\noindent\textbf{Classifier Architecture:}
{The set of positive class image patches (satisfying eq.} (\ref{eq:transPositive}){) and negative class image patches (satisfying eq.} (\ref{eq:transNegative})) are first normalized to $[0,1]$ range and then stacked together to train the framework presented in Sec. \ref{sec:sdae}. As mentioned in Section \ref{sec:data}, the digital resolution in all three datasets used in our experiments are $12$ bits ($4096$ gray-levels, DICOM tag$=\left(0028, 0101\right)$; unsigned, DICOM tag$=\left(0028, 0103\right)$); therefore, the CXR intensities are divided by 4096 to achieve normalization. For training datasets acquired under different protocols, corresponding DICOM tags can be used to decide the normalization. Moreover, in our experiments, $r$ is set to $7$ based on performance accuracy and efficiency ($r=3\dots 13$ was tested); therefore, each training patch has dimensions of $Nx\times(2*r+1)=2048\times 15$ or $(2*r+1)\times Ny=15\times 2048$. The dimension of the multiple layer SdAE is \begin{small}$30720\times 800\times 400$\end{small}. For SdAE, we use the sigmoid activation function, learning rate of 0.001, batch size of 1000, and 100 epochs. For DNN we use the sigmoid activation, batch size of 1000, learning rate of 0.1, and 100 epochs. Again, the parameters of the network are empirically estimated to minimize the reconstruction error. Furthermore, the number of layers has been decided empirically: the layers are added to the network until the reconstruction error stops decreasing. The proposed deep learning architecture for line estimation is shown in Fig. \ref{fig:saeall}a.

\noindent\textbf{Hypothesis Testing:} Each pixel row (or column) along the axis (as shown in Fig. \ref{fig:saeall}a) is tested for the line position using the trained classifiers. Similar to the practice adopted in \cite{zheng2008four}, the position of the line is determined by averaging the top candidates (10 in our experiments) with the highest score in order to make the framework robust to classification noise. The four intersecting vertices of bounding lines are used to extract $\widehat{\mathbf{T}}$ and $\widehat{\mathbf{S}}$ using a sequence of well-known geometrical operations.

\subsubsection{Orientation Estimation}
\noindent\textbf{Training:} The orientation estimation hypothesis in ESL is formulated as: finding the object of interest with centroid at position $\mathbf{T}$ having anisotropic scale $\mathbf{S}$ and orientation $\theta$. Using a bounding box with position and anisotropic scale already estimated, the hypotheses for orientation estimation is generated by rotating the bounding box of size ($\mathbf{S}$) around ($\mathbf{T}$). The position-scale-orientation (anisotropic similarity) hypothesis is positive if, in addition to satisfying (\ref{eq:transPositive}) for all four lines, it also satisfies:
\begin{small}$|\theta-\widehat{\theta}|\le \Delta\theta^{+} \text{rad}$\end{small}, where $\theta$ denotes the orientation of the bounding box encapsulating the ground-truth label and $\widehat{\theta}$ is the hypothesized orientation. A negative hypothesis satisfies \begin{small}$|\theta-\widehat{\theta}|\ge \Delta\theta^{-} \text{rad}$\end{small} (\begin{small}$\Delta\theta^{+} \ne \Delta\theta^{-}$\end{small}). In our experiments, we use $\Delta\theta^{+}=0.017$ rad (1 degree) and $\Delta\theta^{-}=0.034$ rad.

\begin{figure}
\centering
\begin{subfigure}[b]{0.45\textwidth}
\includegraphics[width=\textwidth]{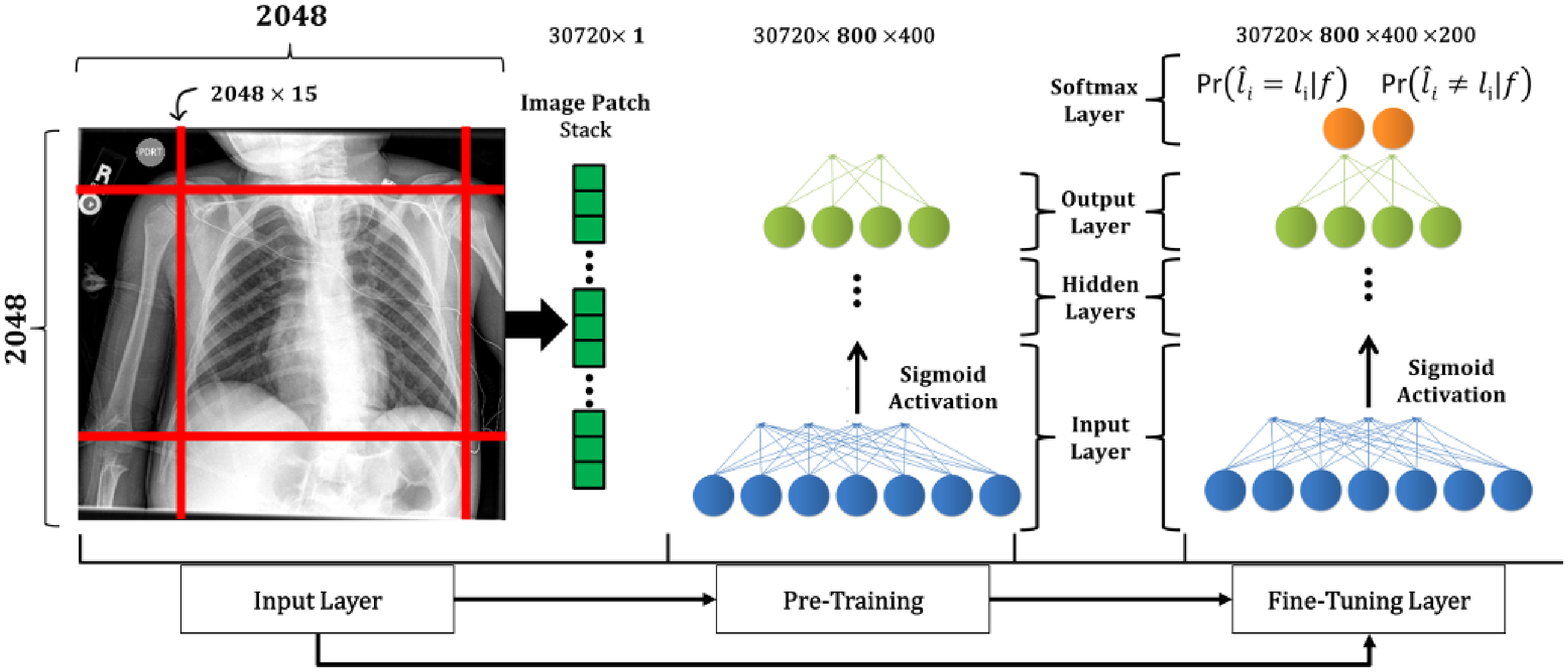}
\caption{}
\end{subfigure}
\begin{subfigure}[b]{0.45\textwidth}
\includegraphics[width=\textwidth]{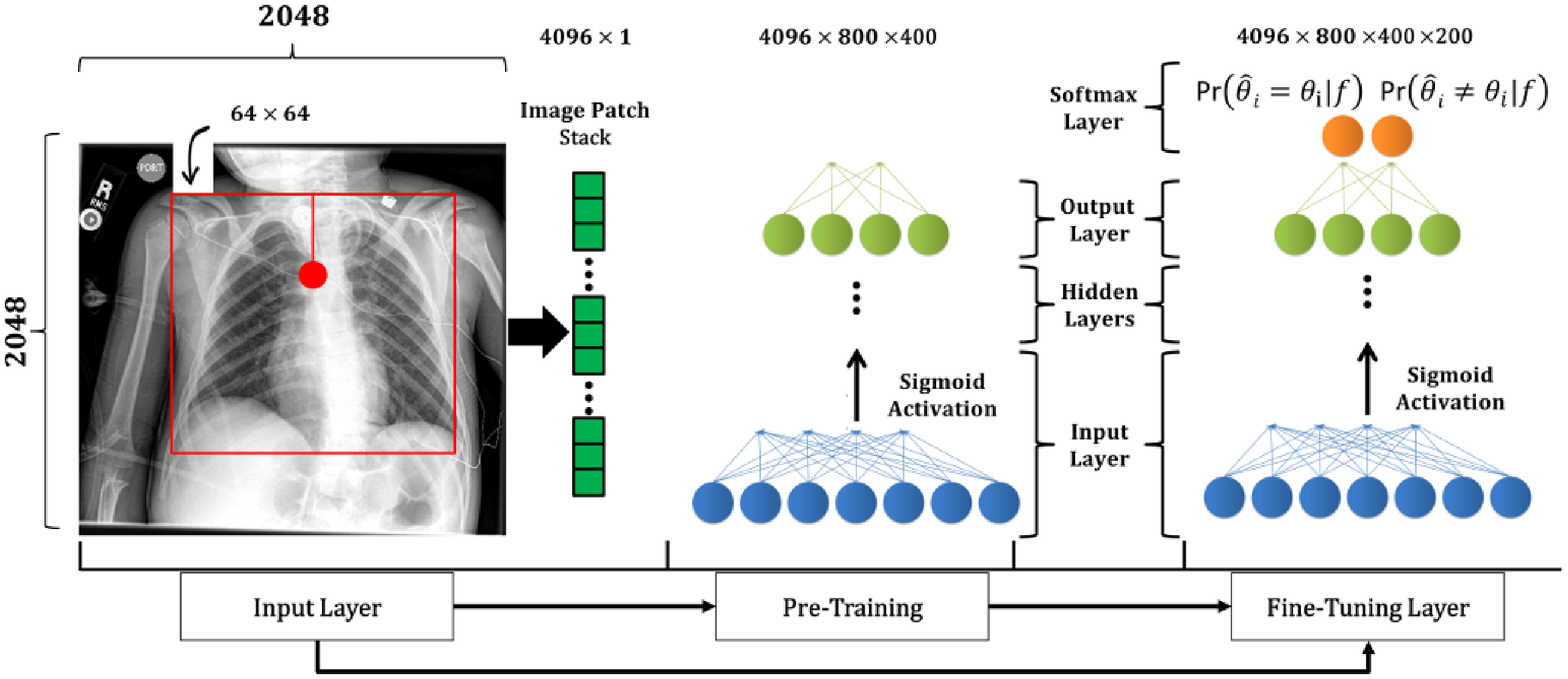}
\caption{}
\end{subfigure}
\includegraphics[width=0.43\textwidth]{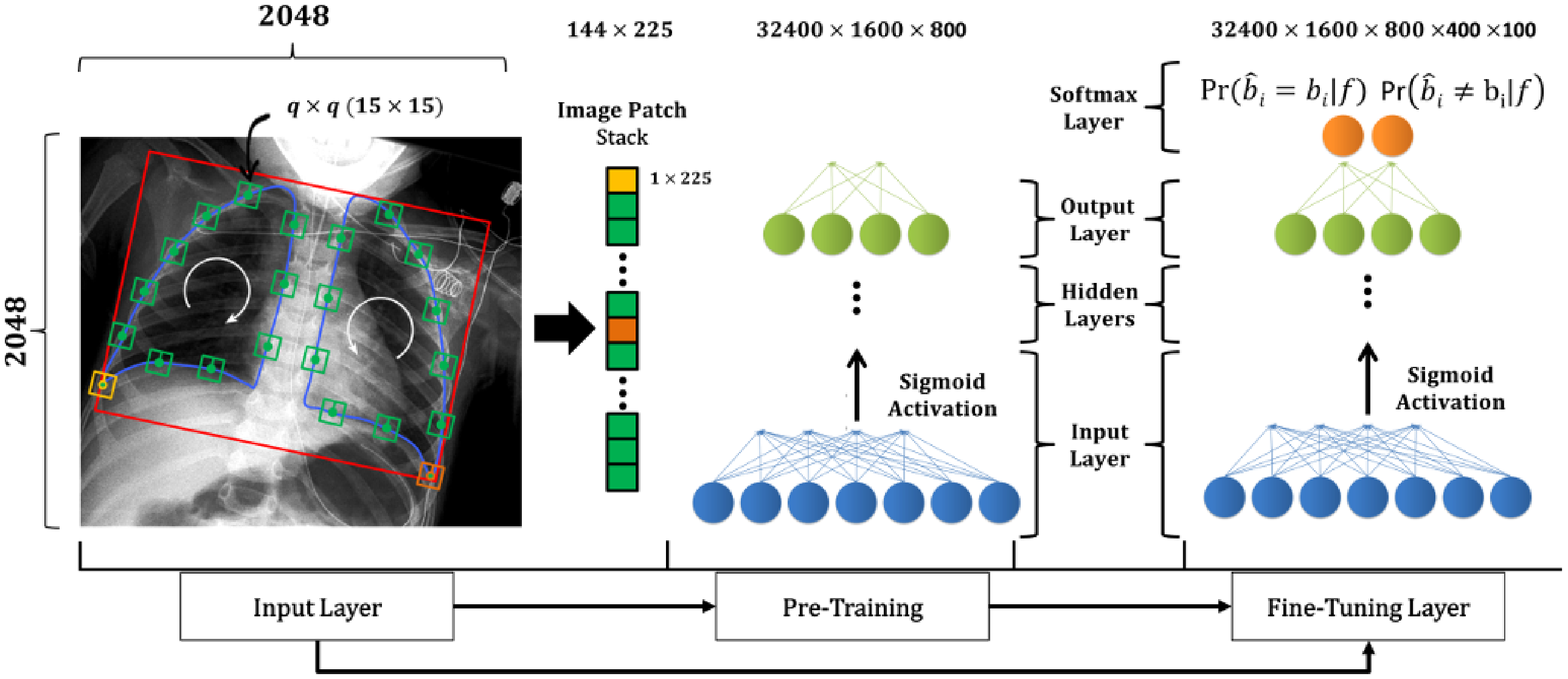}
\begin{subfigure}[b]{\textwidth}
\caption{}
\end{subfigure}
\caption{\footnotesize{Architectures of the proposed deep learning framework for (a) line estimation, (b) orientation estimation, the pendulum indicates the orientation of the bounding box and (c) modes of shape variation. The squares indicate the patches (size = $q\times q$) extracted around landmarks. Yellow and orange patches indicate the first patch for right and left lungs respectively. The clock-wise and counter clock-wise white arrows indicate the direction of concatenation of patches to obtain the feature set.}}
\label{fig:saeall}
\end{figure}

\noindent\textbf{Classifier Architecture:}
For computational efficiency and feature uniformity, the extracted patches using the oriented bounding box are resized to \begin{small}$64\times 64$\end{small} pixels using B-spline interpolation. The proposed architecture and its configuration for orientation estimation are shown in Fig. \ref{fig:saeall}b. The architecture for SdAE and DNN uses same hyper parameters as the bounding line classifiers (Section \ref{sec:edge}).

\noindent\textbf{Hypothesis Testing:} A bounding box with estimated position and anisotropic scale from Section \ref{sec:edge} is
rotated with a step-size of $0.0017$ radians. Subsequently, trained orientation classifier is used to calculate the similarity scores for each rotated hypothesis. The final estimate is obtained using the average of top 10 candidates. 

\subsection{Optimal Mean Shape Determination}
\label{sec:meanshape}
For optimal performance using SSM-based segmentation methods, the mean contour shape has to be initiated as close to the true boundary as possible \cite{tecootes1994active}. As the anatomical structure of the lung evolves with age, resulting in shape variation amongst various age groups; therefore, we evaluate a multiple shape modeling approach for our generic framework. Based on maximum likelihood estimation of the Gaussian mixture model (GMM) clustering of the aspect-ratios (\begin{small}${\widehat{\mathbf{S}}_x}/{\widehat{\mathbf{S}}_y}$\end{small}; Fig. \ref{fig:adultPedsAspectRatio}); the optimal number of shape models for our dataset is determined to be two for our training dataset. The aspect-ratio is also found to be strongly correlated with the shape variation of modes weighted by eigenvalues (\begin{small}R=$-0.945$\end{small}). Therefore, the training data is partitioned into two groups based on the aspect-ratio. 

\noindent\textbf{Training:} Let \begin{small}$\{\mathbf{x}_n^{(i)}\}_{n=1}^{N_i}$\end{small} define a set of $N_i$ training shapes for the group $i\in \{1, 2\}$ then the optimal mean shape \begin{small}$\overline{\mathbf{x}}_\text{opt}^{(i)}$\end{small} for that group are obtained iteratively by minimizing the following residual error after generalized Procrustes alignment of $N_i$ group training shapes,

\begin{small}
\vspace{-0.15in}
\begin{eqnarray}
\widehat{\overline{\mathbf{x}}}_\text{opt}^{(i)} = \mathop {{\rm{argmin}}}\limits_{\overline{\mathbf{x}}^{(i)}} \sum\limits_{n = 1}^{N_i} {\sum\limits_{m = 1}^M {{{\left\| {\mathcal{T}_n(\overline{\mathbf{x}}(m)^{(i)}) - {\mathbf{x}(m)_n^{(i)}}} \right\|}^2}} },
\label{eq:meanshape}
\end{eqnarray}
\end{small}
where $\mathcal{T}_n()$ denotes the generalized Procrustes transformation from the mean shape $\overline{\mathbf{x}}^{(i)}$ to a training shape $\mathbf{x}_n^{(i)}$.

\noindent\textbf{Hypothesis Testing:} The appropriate shape model for the test image is chosen based on the estimated aspect-ratio (\begin{small}${\widehat{\mathbf{S}}_x}/{\widehat{\mathbf{S}}_y}$\end{small}). A threshold of $1.22$ for the aspect-ratio is empirically determined to decide between appropriate shape model for the test image.

\subsection{Shape Parameter Estimation}
\label{sec:mashdl}
\begin{figure}
\includegraphics[width=0.5\textwidth]{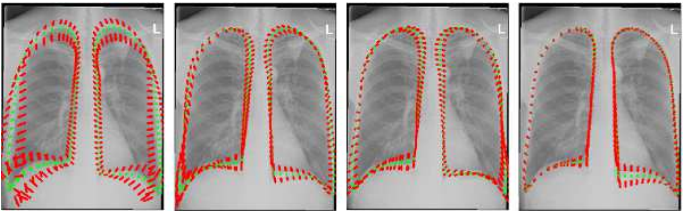}
\caption{\footnotesize{{Four highest modes of variation in training data are shown from left to right. For each mode, we show the superimposed synthesized shapes (step size}$=0.05$): {The landmarks forming the positive shape hypotheses of the mode is shown in green while the landmarks forming the negative shape hypotheses} (eq. (\ref{eq:modesNegative})) are show in red.}}
\label{fig:modes}
\end{figure}

The concept of using representation learning methods for SSM is not novel. A few attempts have already been made in the literature such as \cite{zheng2008four} where irregular sampling patterns were used to capture the shape deformation followed by Haar-wavelet feature extraction. However, the need for extracting optimal hand-crafted features, the amount of training data needed to learn shape parameters simultaneously, as well as the computational complexity of the multi-parameter classifier made representation learning methods a less attractive choice compared to the conventional iterative optimization techniques. Our proposed approach, Marginal Shape Deep Learning (MaShDL), attempts to address these challenges. To learn $\Omega_\text{Shape}$, MaShDL adopts a recursive rather than an iterative approach adopted by the conventional SSM (eq. (\ref{eq:modeweights})) \cite{cootes1995active}. Specifically, instead of estimating and optimizing all $K$ modes collectively (eq. (\ref{eq:asm})); MaShDL refines the aligned mean shape by recursively adding finer modes. This modification simplifies the hypothesis space by letting separate classifiers trained for each mode. From eq. (\ref{eq:asm}), (\ref{eq:modeweights}), and (\ref{eq:nonrigid}), the estimated aligned shape $\mathbf{x}_{k}$ using the $k$ largest modes can be recursively written in terms of the aligned shape $\mathbf{x}_{k-1}$, obtained using the $(k-1)$ largest modes,\\
\begin{minipage}{.24\textwidth}
\begin{small}
\begin{eqnarray}
\mathbf{x}_{k}&=&\mathbf{x}_{k-1}+\mathbf{p}_k\mathbf{b}_k,\nonumber
\end{eqnarray}
\end{small}
\end{minipage}
\begin{minipage}{.24\textwidth}
\begin{small}
\begin{eqnarray}
\mathbf{x}_0&=&\overline{\mathbf{x}}^{(i)},
\label{eq:ssmsequential}
\end{eqnarray}
\end{small}
\end{minipage}

\noindent where $\mathbf{p}_k$ is the $k^\text{th}$ eigenvector and $\mathbf{b}_k$ is the corresponding weight. It is important to mention here that modes and mean shape in eq. (\ref{eq:ssmsequential}) are based on the grouping performed in section (\ref{sec:meanshape}), the superscript $(i)$ is dropped for the ease of reading. Eq. (\ref{eq:ssmsequential}) transforms (\ref{eq:asm}) from block parameter estimation of the modes (as performed in eq. (\ref{eq:modeweights})) to recursive estimation by successively adding the next lower order mode. Moreover, eq. (\ref{eq:nonrigid}) and (\ref{eq:ssmsequential}) imply that $\Omega_\text{Shape}$ is nowhere commutative subgroup since the $k^\text{th}$ largest mode of variation has to be estimated prior to $(k+1)^\text{th}$ mode. Therefore, parameter estimation through representation learning in MaShDL starts with the most informative (highest) mode and sequentially adds lower variability modes.\\
\noindent\textbf{Training:} MaShDL begins by learning the highest mode through deforming the mean shape (or the zeroth mode: $\mathbf{x}_0$) that is subsequently deformed by the second highest mode and so on.\\
\noindent\underline{Positive Shape Hypotheses:} The positive shape hypothesis for all modes are the same. The positive hypothesis corresponds to extracting $q\times q=15\times 15$ patches around these $M$ landmarks (=144 (72 per lung)) of manually delineated ground truth shape. \\
\noindent\underline{Negative Shape Hypotheses:} The negative hypotheses for the $k^\text{th}$ mode are fabricated as follows: 
\begin{tasks}(1)
\task Use eq. (\ref{eq:asm}) and the mean shape $\overline{\mathbf{x}}^{(i)}$ to estimate the $K$ ``true'' modes of variation of $n_i^\text{th}$ shape in the training set. 
\task To generate negative hypotheses for the $k^\text{th}$ mode of the $n_i^\text{th}$ shape, generate a set of synthesized shapes by keeping the \begin{small}$(k-1)$\end{small} largest estimated modes from (a) constant and varying only the $k^\text{th}$ mode within the range (\begin{small}$\widehat{\mathbf{b}}_k\le|3\lambda_k|$\end{small}). Henceforth, a negative hypothesis should satisfy eq. (\ref{eq:rigid}) and

\begin{small}
\vspace{-0.1in}
\begin{eqnarray}
\widehat{\mathbf{x}}_{k}&=&\widehat{\mathbf{x}}_{k-1}+\mathbf{p}_k\widehat{\mathbf{b}}_k\nonumber\\
\left| {\mathbf{b}_k - \widehat{\mathbf{b}}_k} \right|&\ge&0.25, \forall~\widehat{\mathbf{b}}_k\in[-3.0,\dots,+3.0]\lambda_k
\label{eq:modesNegative}
\end{eqnarray} 
\end{small}
\noindent$\mathbf{b}_k$ is the value of the $k^\text{th}$ mode obtained using eq. (\ref{eq:modeweights}). $\left| {\mathbf{b}_k - \widehat{\mathbf{b}}_k} \right|\ge 0.25$ translates, in our application, to a minimum landmark-to-landmark distance of 2 pixels. 
\task Extracting $q\times q$ patches (shown as squares in Fig. \ref{fig:saeall}c) around $M$ points of the shape synthesized.
\end{tasks}
Fig. \ref{fig:modes} shows examples of positive (in green) and negative hypotheses (in red) for the four highest modes. Each hypothesis corresponds to a shape depicted by the concatenation of patches of size $q\times q$, each extracted around the $M$ (red or green) landmark points. The extracted hypotheses are subsequently used to train a classifier for the $k^\text{th}$ deformation mode. Similar to conventional SSM, our framework uses local appearance information to move the object boundary to the optimal position. 

\noindent\textbf{Classifier Architecture:}
In our experiments, identical patch sizes (\begin{small}$q=15$\end{small}) are used for training classifiers for all modes (\begin{small}$q=3\dots 21$\end{small} were tested). Smaller patch sizes are found to be prone to noise while the higher sizes tend to miss subtle shape deformations. The image patches extracted around every landmark point are subsequently stacked together in a specific order (Fig. \ref{fig:saeall}c) to form a single hypothesis. Each training hypothesis has dimensions of \begin{small}$172\times 225$\end{small} pixels. A multiple layer (\begin{small}$38700\times 1600\times 800$\end{small}) SdAE followed by DNN is used (shown in Fig. \ref{fig:saeall}c). For SdAE, sigmoid activation function, learning rate of 0.001, batch size of 1000, and 100 epochs are used.\\
\noindent\textbf{Hypothesis Testing:}
The optimal mean shape (obtained in Section \ref{sec:meanshape}) is first aligned to the detected object in the test image using $\widehat{\Omega}_\text{Space}$. Next, the trained classifier for the largest mode of shape variation is used to deform the aligned mean shape $\overline{\mathbf{X}}$ followed by classifier trained for the second highest mode and so on. The process is iterated to estimate the next highest variation mode until a cumulative energy of $95\%$, which, in our application, is equivalent to using the largest fifteen modes of variation are included. Limiting the number of modes is a common practice when creating PCA-based statistical shape models \cite{cootes1995active}. {Although there is no theoretical limitation on learning all $M$ modes using MaShDL, a larger training dataset is generally needed to train classifiers for lower-ranked modes due to the increasing subtlety between the positive and negative hypotheses with the number of modes. Moreover, it is also predicted that both the total number of estimate-able modes as well as the machine-discernibility of adjacent modes are correlated with both digital and spatial image resolution.}

\subsection{Data Augmentation}
Since the number of positive hypotheses in our training routine is smaller than the number of negative hypotheses, a data augmentation approach similar to the one presented in \cite{krizhevsky2012imagenet} along with random sampling is adopted to balance the samples prior to training. Specifically, we used two forms of data augmentation: (1) geometrical augmentation, and (2) appearance augmentation. The geometrical augmentation consists of generating horizontal and vertical reflections of hypotheses while the appearance augmentation consists of slightly altering the intensities in the training images. {For intensity alteration we first perform PCA over the entire training dataset ($\{f_n\}_{n=1}^N$). Subsequently, to the $n^\text{th}$ training normalized image ($0,1$), we add the multiples of the extracted principal component with magnitude proportional to the corresponding eigenvalue times a random variable drawn from a Gaussian distribution with zero mean and 0.1 standard deviation, i.e., \[[\mathbf{p}_1^f, \mathbf{p}_2^f,\dots,\mathbf{p}_N^f][\alpha_1\lambda_1^f, \alpha_2\lambda_2^f,\dots, \alpha_N\lambda_N^f]^T\] where $\mathbf{p}_n^f$and $\lambda_n^f$ denotes the $n^\text{th}$ eigenvector and eigenvalue respectively. The superscript $f$ is added to denote the training image data and to differentiate them from eigenvalues and eigenvectors of training shape data defined in section} \ref{sec:parametrized}. The drawn random variable $\alpha_n\sim\mathcal{N}\left(0,0.1\right)$ is applied to every pixel of the $n^\text{th}$ training image. Same augmentation scheme was applied to the hypotheses for every classifier in our framework.

\section{Experimental Results}
\label{sec:results}
The performance of the proposed framework and its individual modules (ESL, MaShDL) was evaluated using two-fold cross-validation. All three datasets (JSRT, BTP, CNHS) were evenly divided into two sets for training and validation then the results were averaged over the two validation rounds.

\subsection{Space Parameter Estimation: MSL vs. ESL}
\label{sec:mslvesl}
\begin{figure}[ht]
\begin{subfigure}[c]{0.25\textwidth}
\centering
\includegraphics[width=\textwidth]{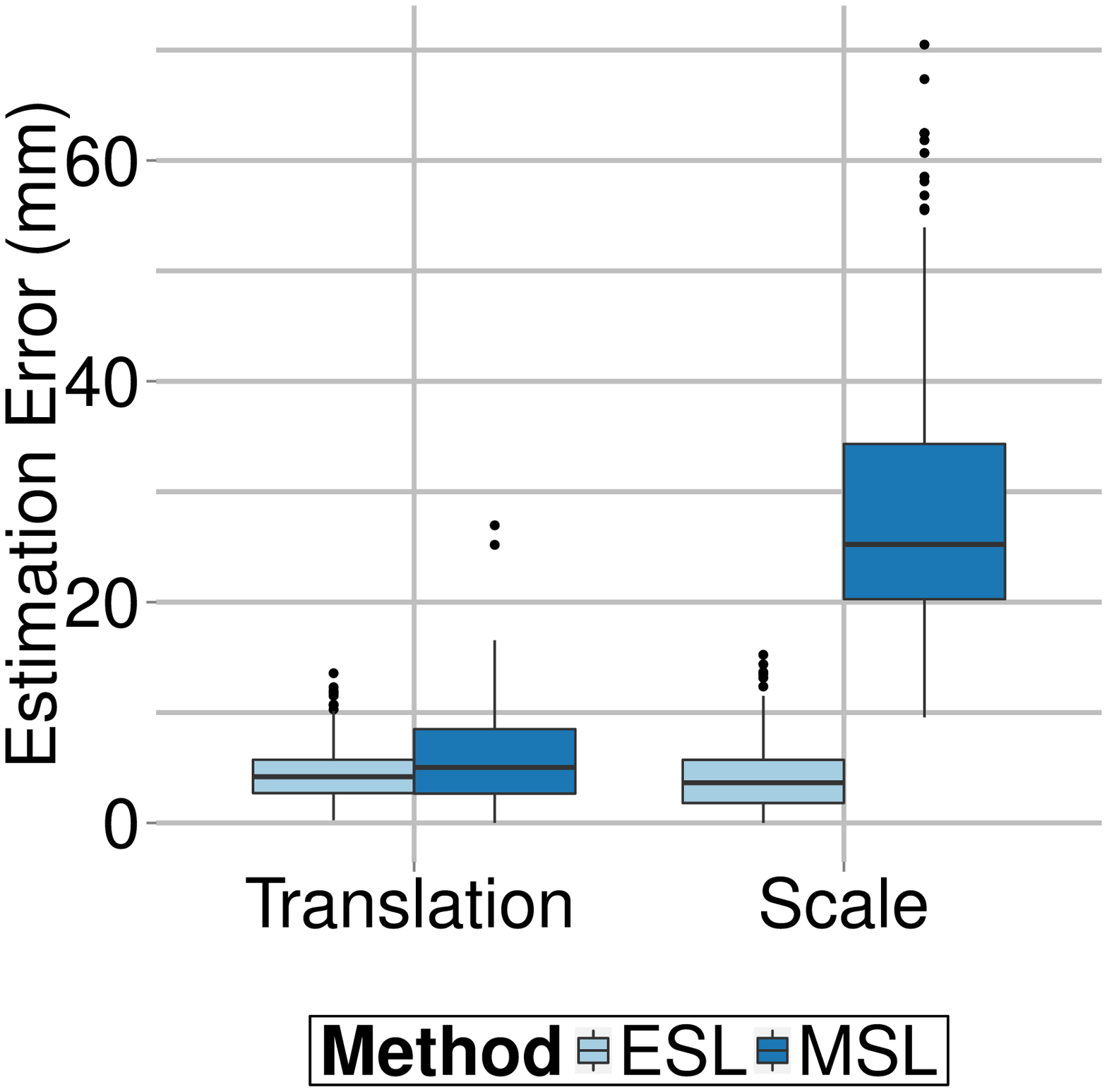}
\caption{}
\label{fig:mslbox}
\end{subfigure}
\begin{subfigure}[c]{0.215\textwidth}
\centering
\includegraphics[width=\textwidth, height=5cm]{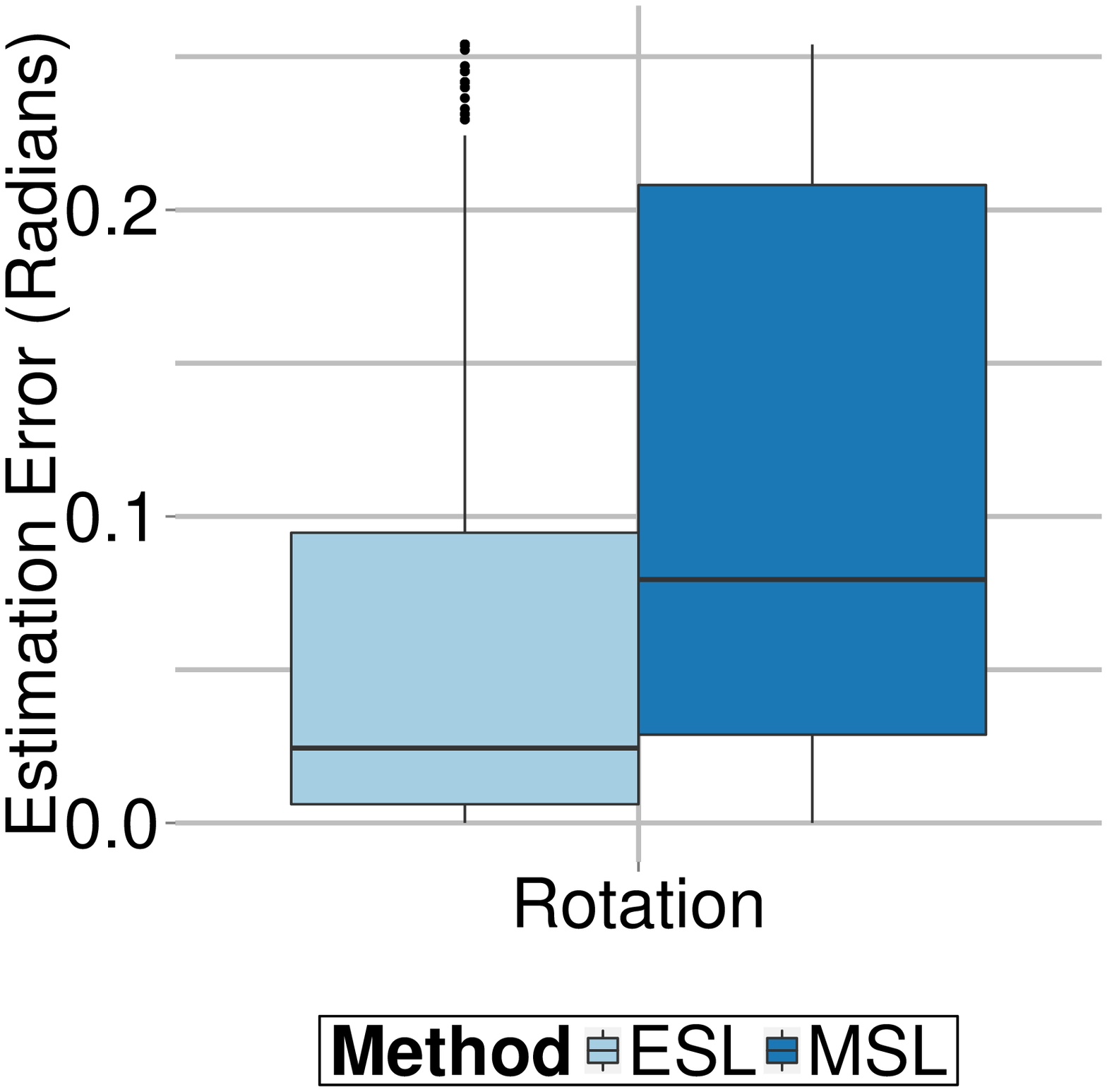}
\caption{}
\label{fig:eslbox}
\end{subfigure}
\caption{\footnotesize{Boxplots of space parameters estimation ($\widehat{{\Omega}}_\text{Space}$) error using MSL \cite{zheng2008four} and ESL (proposed). (a) Translation ($\mathbf{T}$) and scale ($\mathbf{S}$) (b) Orientation $\mathbf{\theta}$.
}}
\label{fig:eslmslbox}
\end{figure}

The performance of ESL and MSL methods were compared using the DL extension of MSL \cite{ghesu2015marginal}. Furthermore, parameters in the original MSL were reordered from eq. (\ref{eq:msl}) for a more meaningful comparison with ESL: translation followed by scale and orientation estimation respectively. Fig. \ref{fig:eslmslbox} presents the estimation error in translation and scale using MSL and ESL. The estimation error in translation was \begin{small}$4.42\pm 2.25$\end{small} mm with the ESL compared to \begin{small}$5.62\pm 3.62$\end{small} mm using MSL ($p$-value\begin{small}$<0.001$\end{small}; Wilcoxon rank sum test). For scale estimation, an average error of \begin{small}$3.99\pm 2.97$\end{small} mm using ESL was obtained compared to \begin{small}$28.09\pm 10.77$\end{small} mm using MSL ($p$-value\begin{small}$<0.001$\end{small}). Although both ESL and MSL follow the same mechanism for orientation estimation, as predicted in eq. (\ref{eq:rigid01}), due to the accumulation of error from translation and scale, MSL achieves an orientation error of \begin{small}$0.11\pm 0.09$\end{small} radians, which was significantly worse than the one obtained using ESL (\begin{small}$0.06\pm0.07$\end{small} radians, $p-$value\begin{small}$<0.001$\end{small}). In our experiments, the average time to perform detection using the ESL pipeline was $5.9$ seconds per CXR, compared $10.3$ seconds for the MSL. Both techniques were implemented in Matlab (The MathWorks, Inc., Natick, MA) and ran using CPU only.

\subsection{Shape Parameter Estimation: MaShDL vs. ASM}
\label{sec:mashdlasm}
\begin{figure}[ht]
\centering
\begin{subfigure}[c]{0.23\textwidth}
\centering
\includegraphics[width=\textwidth]{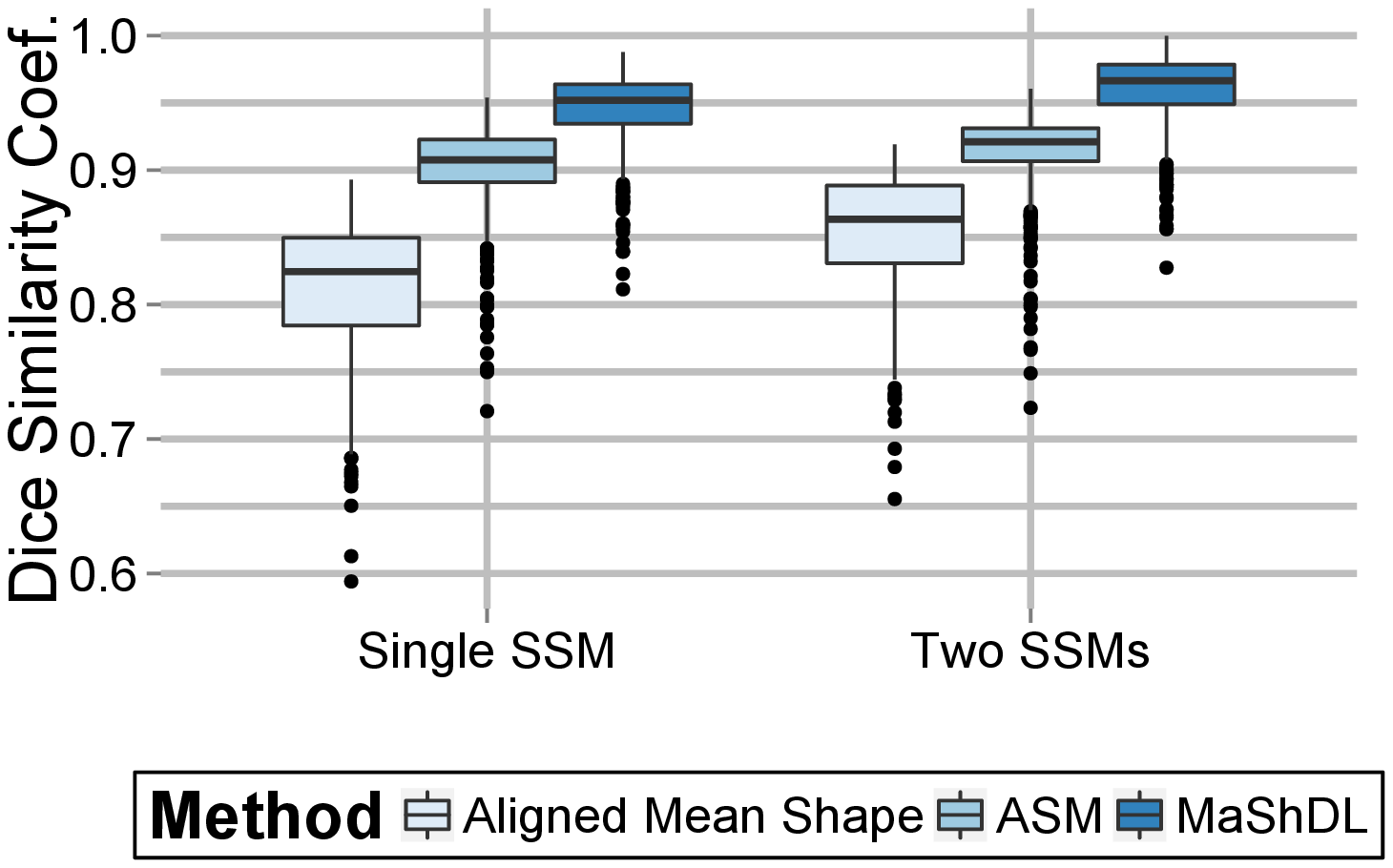}
\end{subfigure}
\begin{subfigure}[c]{0.23\textwidth}
\centering
\includegraphics[width=\textwidth]{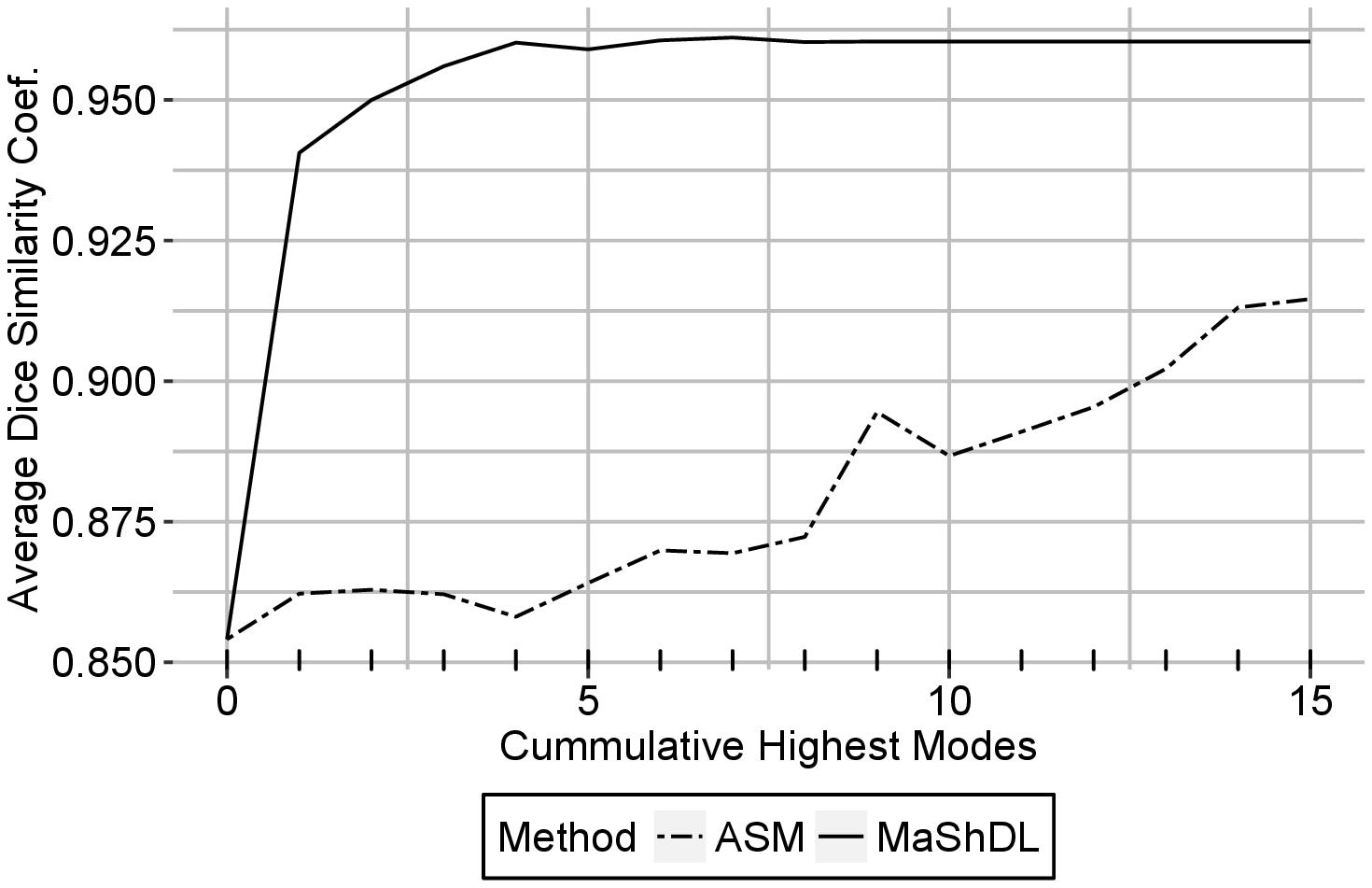}
\end{subfigure}
\caption{\footnotesize{(a) Boxplots of shape parameters estimation ($\widehat{{\Omega}}_\text{Shape}$) error using ASM \cite{cootes1995active} and MaShDL (proposed) using $K=15$ largest modes of shape variation. (b) Average segmentation accuracy, measured using DSC, obtained as a function of the number of shape parameters using ASM \cite{cootes1995active} and MaShDL (proposed). Mode $0$ denotes the aligned mean shape ($\overline{\mathbf{X}}$), the mean shape for both ASM and MaShDL are aligned using ESL.}
}
\label{fig:asmmashdl}
\end{figure}

Fig. \ref{fig:asmmashdl}(a) shows the boxplots of DSC for the lung field segmentation using just the mean shape (baseline), SSM-based ASM \cite{cootes1995active}, and the MaShDL (using single and two SSMs). A single model was created using the training data from all three datasets. Two separate shape models were created using the clustering criteria described in Section \ref{sec:meanshape}. Mean shape initialization was performed using ESL (Section \ref{sec:esl}). The best results were achieved with the two SSMs; however, in both cases, MaShDL significantly outperforms the conventional ASM ($p$-value$<0.001$ for single SSM, $p$-value$<0.001$ for two SSM). A mean DSC of \begin{small}$0.85\pm0.04$\end{small} was obtained using just the mean shape alignment through ESL, \begin{small}$0.92\pm0.03$\end{small} using ASM, and \begin{small}$0.96\pm0.03$\end{small} using MaShDL. The results in the boxplot are reported using the modes carrying \begin{small}$95\%$\end{small} cumulative energy (\begin{small}$K=15$\end{small}).
\begin{table*}[htb]
\centering
\caption{{\color{black}\footnotesize{\uppercase{Quantitative comparison of the proposed approach with the current state-of-the-art methods in lung field segmentation.}}\label{table:methodsPerformance}}}
\begin{scriptsize}
\begin{tabular}{@{\extracolsep{\stretch{3}}}p{2.8cm}ccc}
\toprule
\multirow{2}{*}{\textbf{Method}}&\textbf{Overlap}&\textbf{ACD (mm)}&\textbf{DSC}\\
&\multicolumn{3}{c}{{\color{black}\textbf{Mean$\mathbf{\pm}$Standard Deviation (Min/Max)}}}\\
\midrule
\cite{annangi2010region}&-&-&$0.880\pm0.070$\\
\cite{dawoud2010fusing} $\dagger$&$0.940\pm0.005$&$2.460\pm2.060$&-\\
\cite{sohn2011segmentation} $\dagger$&$0.851\pm0.046$&-&$0.952\pm0.016$\\
\cite{shi2008segmenting} $\ddagger$&$0.920\pm0.031$&$2.492\pm1.092$&-\\
\cite{van2000automatic}\\
\hspace{0.25cm} Pixel Classification (PC)&$0.938\pm0.027$&$3.250\pm2.650$\\
\hspace{0.25cm} AAM Whiskers&$0.913\pm0.032$&$2.700\pm1.100$\\
\hspace{0.25cm} ASM tuned&$0.927\pm0.032$&$2.300\pm1.030$\\
\hspace{0.25cm} Hybrid AAM+PC&$0.933\pm0.025$&$2.060\pm0.840$&-\\
\hspace{0.25cm} Hybrid ASM+PC&$0.934\pm0.037$&$2.080\pm1.400$&-\\
\hspace{0.25cm} PC+Post-processing&$0.945\pm0.022$&$1.610\pm0.800$&-\\
\hspace{0.25cm} Hybrid Voting&$0.949\pm0.020$&$1.620\pm0.660$&-\\
\cite{shao2014hierarchical} $\dagger$&$0.946\pm0.019$&$1.669\pm0.762$&$0.972\pm0.010$\\
\cite{candemir2014lung} $\ddagger$&$0.954\pm0.015$&$1.321\pm0.316$&-\\
{\color{black}\cite{ibragimov2016accurate}} $\dagger$&{\color{black}$0.953\pm0.2$}&{\color{black}$1.430\pm0.850$}&-\\
{\color{black}\cite{dai2017scan}} $\dagger$&{\color{black}$0.929\pm0.500$}&-&{\color{black}$0.963\pm0.3$}\\
\midrule
\multicolumn{4}{c}{\textbf{Inter-Observer Agreement}}\\
\\
With Retro-Cardiac Region&$0.932\pm0.033$&$1.726\pm 1.252$&$0.951\pm 0.029$\\
Without Retro-Cardiac Region \cite{van2006segmentation}&$0.946\pm 0.018$&$1.640\pm 0.690$&-\\
\midrule
\multicolumn{4}{c}{\textbf{{\color{black}U-Net} \cite{wang2017segmentation} {\color{black}(With Retro-Cardiac Region)}}}\\
\\
\hspace{0.25cm} Overall&$0.91\pm 0.059 \left(0.603/0.971\right)$&$1.622\pm 0.826 \left(0.540/4.872\right)$&$0.941\pm 0.065 \left(0.611/0.985\right)$\\
\hspace{0.25cm} JSRT&$0.91\pm 0.039 \left(0.678/0.960\right)$&$2.974\pm 0.819 \left(1.361/4.959\right)$&$0.952\pm 0.023 \left(0.808/0.980\right)$\\
\hspace{0.25cm} BTP&$0.935\pm 0.031 \left(0.837/0.973\right)$&$2.689\pm 0.563 \left(2.126/3.251\right)$&$0.966\pm 0.017 \left(0.912/0.986\right)$\\
\hspace{0.25cm} CNHS&$0.897\pm 0.068 \left(0.607/0.967\right)$&$1.283\pm 0.603 \left(0.459/4.981\right)$&$0.927\pm 0.078 \left(0.714/0.983\right)$\\
\midrule
\multicolumn{4}{c}{\textbf{{\color{black}U-Net} \cite{wang2017segmentation} {\color{black}(Without Retro-Cardiac Region)}}}\\
\\
\hspace{0.25cm} Overall&$0.904\pm 0.063 \left(0.631/0.980\right)$&$1.237\pm 0.775 \left(0.454/3.836\right)$&$0.948\pm 0.037 \left(0.774/0.990\right)$\\
\hspace{0.25cm} JSRT&$0.949\pm 0.029 \left(0.759/0.976\right)$&$1.775\pm 1.252 \left(0.723/3.921\right)$&$0.973\pm 0.016 \left(0.863/0.988\right)$\\
\hspace{0.25cm} BTP&$0.842\pm 0.075 \left(0.643/0.937\right)$&$2.218\pm 0.426 \left(1.891/3.156\right)$&$0.912\pm 0.046 \left(0.783/0.967\right)$\\
\hspace{0.25cm} CNHS&$0.877\pm 0.055 \left(0.600/0.965\right)$&$1.343\pm 0.747 \left(0.514/7.905\right)$&$0.933\pm 0.034 \left(0.747/0.976\right)$\\
\midrule
\multicolumn{4}{c}{\textbf{Proposed Method (With Retro-Cardiac Region)}}\\
\\
\hspace{0.25cm} Overall&$0.942\pm0.063 \left(0.798/0.972\right)$&$1.536\pm0.685 \left(0.414/2.433\right)$&$0.961\pm0.026 \left(0.788/0.989\right)$\\
\hspace{0.25cm} JSRT&$0.949\pm0.041 \left(0.821/0.972\right)$&$1.878\pm0.708 \left(0.414/2.711\right)$&$0.950\pm0.016 \left(0.841/0.989\right)$\\
\hspace{0.25cm} {BTP}&$0.945\pm0.087 \left(0.801/0.968\right)$&$1.471\pm 0.429 \left(0.765/2.364\right)$&$0.963\pm0.023 \left(0.828/0.987\right)$\\
\hspace{0.25cm} {CNHS}&$0.939\pm0.101 \left(0.798/0.957\right)$&$1.466\pm0.744 \left(0.726/2.433\right)$&$0.946\pm0.032 \left(0.788/0.982\right)$\\
\midrule
\multicolumn{4}{c}{\textbf{Proposed Method (Without Retro-Cardiac Region)}}\\
\\
\hspace{0.25cm} Overall&$0.954\pm0.082 \left(0.771/0.979\right)$&$1.492\pm0.788 \left(0.691/2.786\right)$&$0.969\pm0.038 \left(0.764/0.988\right)$\\
\hspace{0.25cm} JSRT&$0.958\pm0.066 \left(0.844/0.974\right)$&$1.662\pm0.881 \left(0.691/2.548\right)$&$0.969\pm0.019 \left(0.852/0.988\right)$\\
\hspace{0.25cm} BTP&$0.957\pm0.091 \left(0.771/0.979\right)$&$1.616\pm 0.494 \left(0.808/2.787\right)$&$0.972\pm0.037 \left(0.764/0.988\right)$\\
\hspace{0.25cm} CNHS&$0.948\pm0.093 \left(0.822/0.967\right)$&$1.315\pm0.817 \left(0.479/2.234\right)$&$0.967\pm0.053 \left(0.813/0.979\right)$\\
\midrule\\
\end{tabular}\vspace{-0.1in}
\begin{tabular}{ccc}
$\text{DSC}=2\frac{|\text{GT}\cap \text{SEG}|}{|\text{GT}|\cup|\text{SEG}|}$ & $\text{Overlap (Jaccard index)}=\frac{|\text{GT}\cap \text{SEG}|}{|\text{GT}\cup\text{SEG}|}$ &
$\text{ACD}=\frac{1}{2}\left( {\frac{{\sum\limits_{p \in \text{SEG}} {d\left( {p,\text{GT}} \right)} }}{{\left| {\text{SEG}} \right|}} + \frac{{\sum\limits_{p \in \text{SEG}} {d\left( {p,\text{SEG}} \right)} }}{{\left| {\text{GT}} \right|}}} \right)$
\end{tabular}
\begin{tabular}{@{\extracolsep{\stretch{3.0}}}p{11.5cm}}
GT=binary labels of manual ground truth.\\
SEG=binary labels produced by the proposed method.\\
The operator $|.|$ denotes cardinality.\\
{$\dagger$ Method tested on the JSRT dataset.}\\
{$\ddagger$ Method tested on the JSRT dataset among others.}\\
\end{tabular}
\end{scriptsize}
\end{table*}

Fig. \ref{fig:asmmashdl}(b) shows the performance of ASM and MaShDL as a function of cumulative modes of variation (two SSMs). The DL mechanism adopted by MaShDL to extract the local appearance features deforms the shape contour to the true object boundary using less number of modes than ASM. Also from eq. (\ref{eq:modeweights}) and (\ref{eq:ssmsequential}), each atomic unit within ASM and MaShDL have the same order of computational complexity; therefore, MaShDL was demonstrated to be faster than ASM in our experiments for a given performance accuracy. In our experiments, MaShDL framework was also found to be at least four times faster on average than SSM for a given accuracy.

\subsection{Quantitative Comparison with State-of-the-Art Methods}
\begin{figure}
\centering
\begin{subfigure}[b]{0.15\textwidth}
\includegraphics[width=\textwidth]{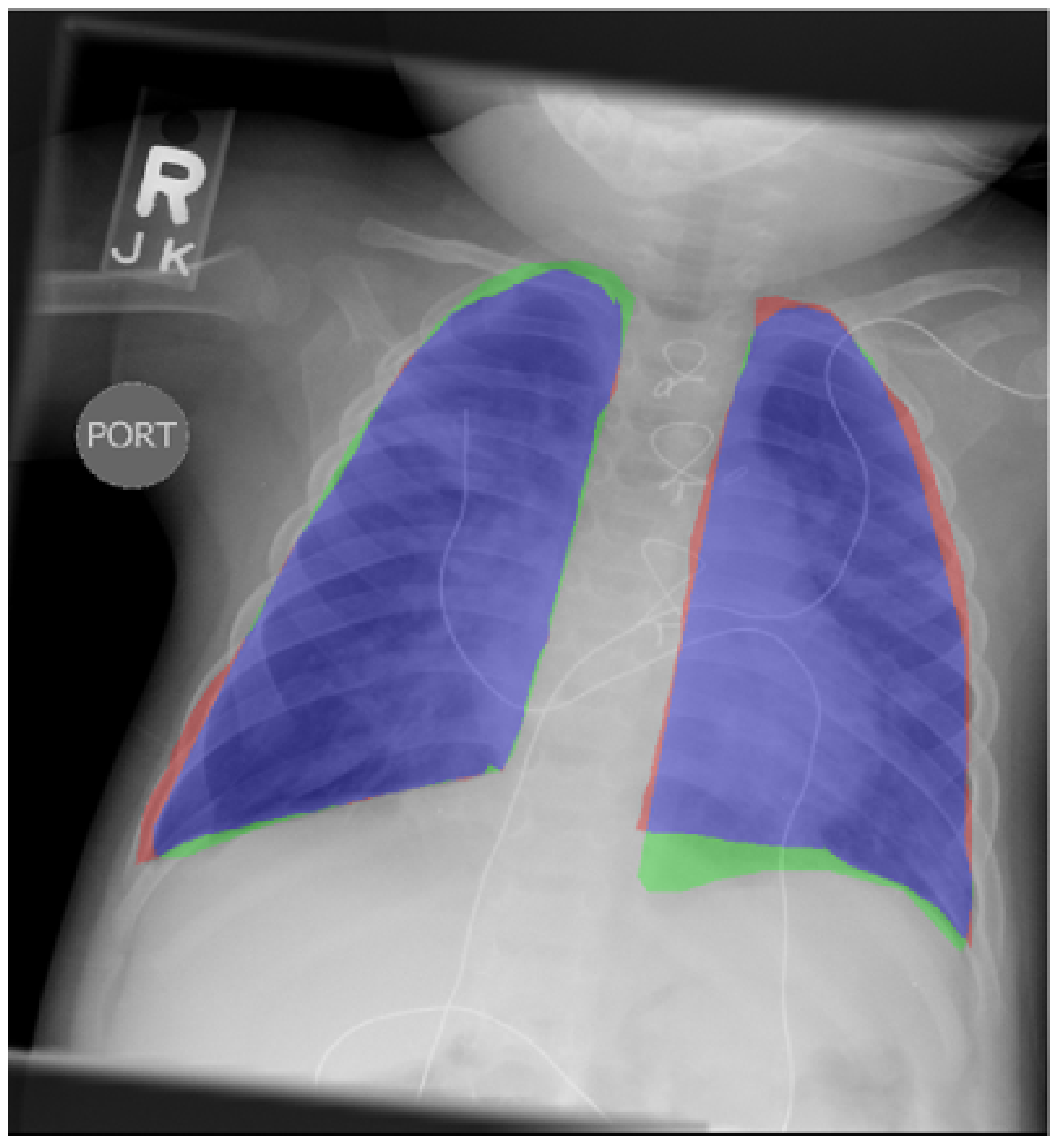}
\caption{\footnotesize{DSC=0.946}}
\end{subfigure}
\begin{subfigure}[b]{0.15\textwidth}
\includegraphics[width=\textwidth]{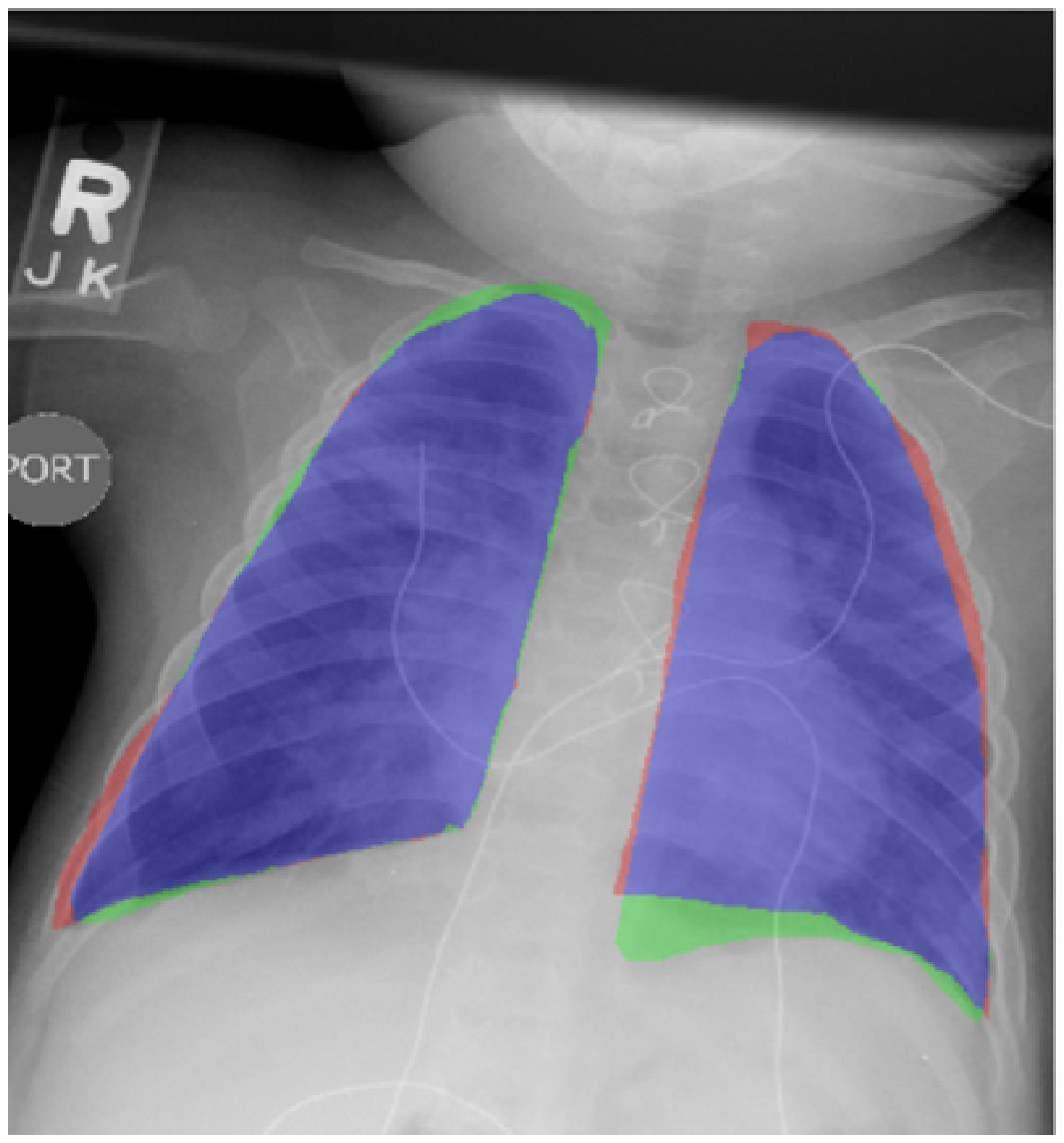}
\caption{\footnotesize{DSC=0.952}}
\end{subfigure}
\begin{subfigure}[b]{0.15\textwidth}
\includegraphics[width=\textwidth]{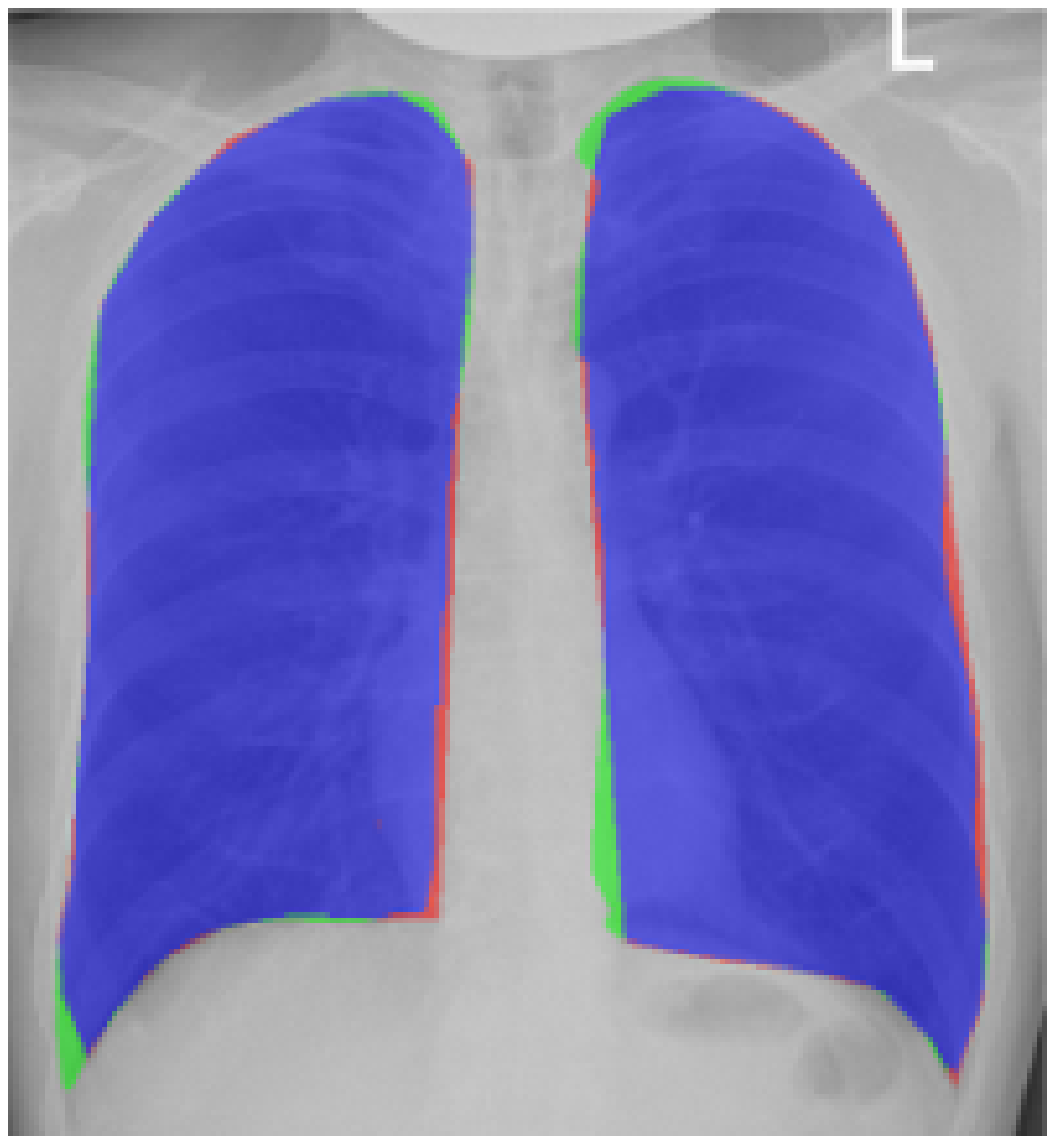}
\caption{\footnotesize{DSC=0.980}}
\end{subfigure}
\caption{\footnotesize{Qualitative lung field segmentation results using the proposed framework (ESL+MaShDL). The cases shown were randomly chosen from the dataset. The segmentation labels obtained are overlaid in over the input CXR. The blue region denotes the overlap area between the ground expert segmented manual ground truth and the segmentation produced using the proposed framework, the green region denotes ground truth area, and the red region denotes the segmentation obtained using the proposed framework. The heart is also visible underneath the segmentation label.}}
\label{fig:qualitative}
\end{figure}

\begin{figure}
\centering
\begin{subfigure}[b]{0.16\textwidth}
\includegraphics[width=\textwidth]{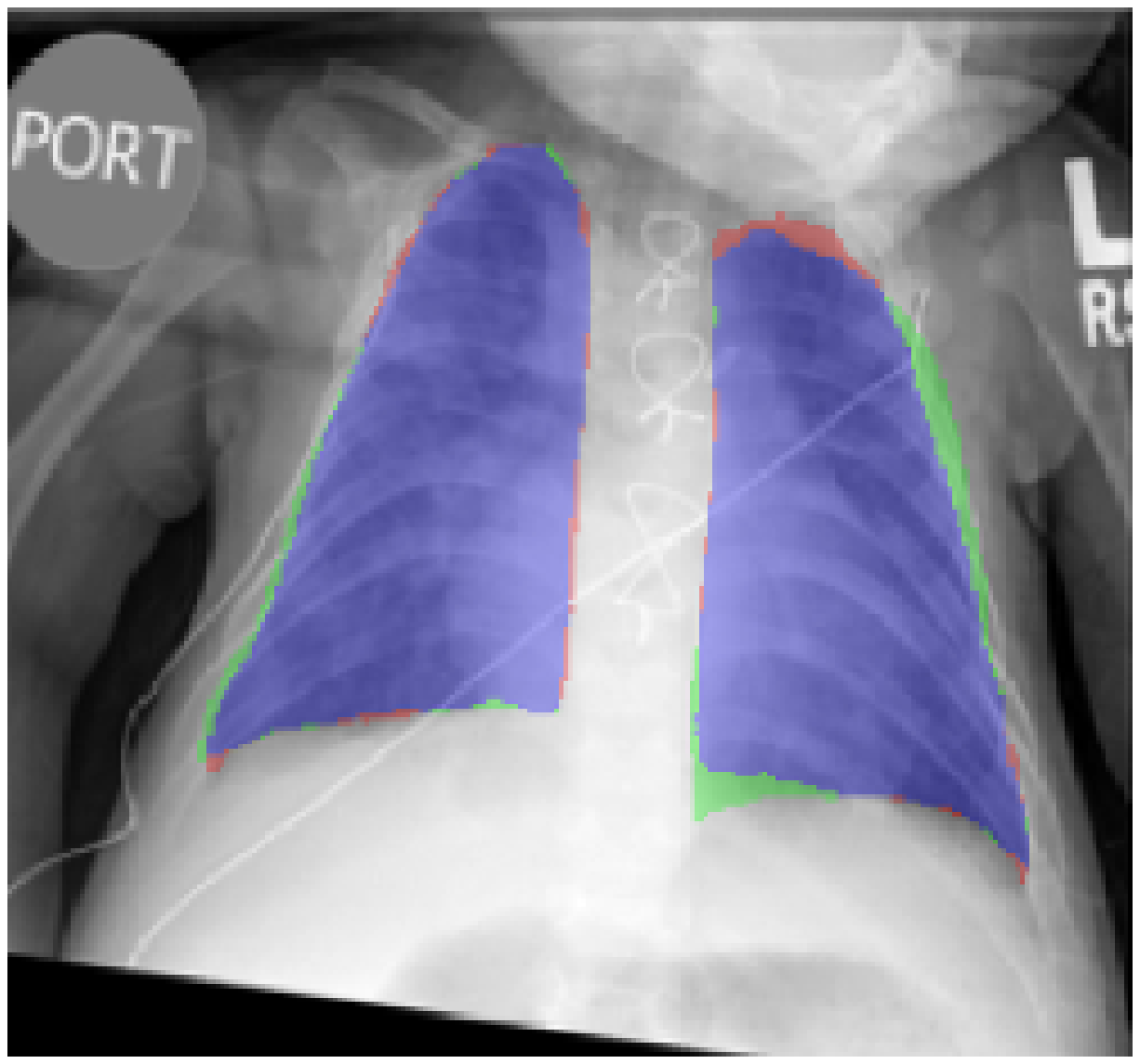}
\caption{\footnotesize{DSC=0.959}}
\end{subfigure}
\begin{subfigure}[b]{0.15\textwidth}
\includegraphics[width=\textwidth]{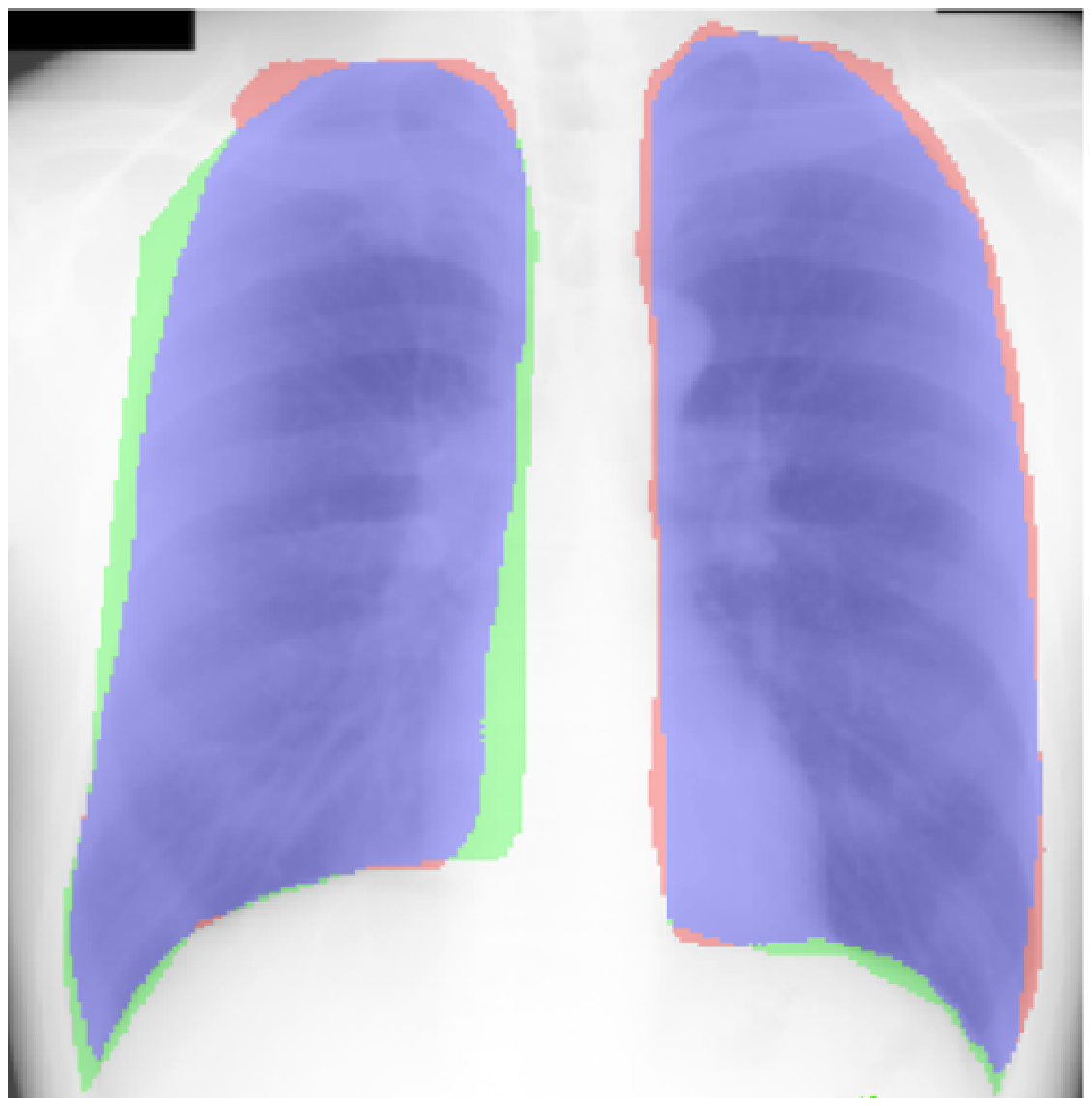}
\caption{\footnotesize{DSC=0.950}}
\end{subfigure}
\begin{subfigure}[b]{0.16\textwidth}
\includegraphics[width=\textwidth]{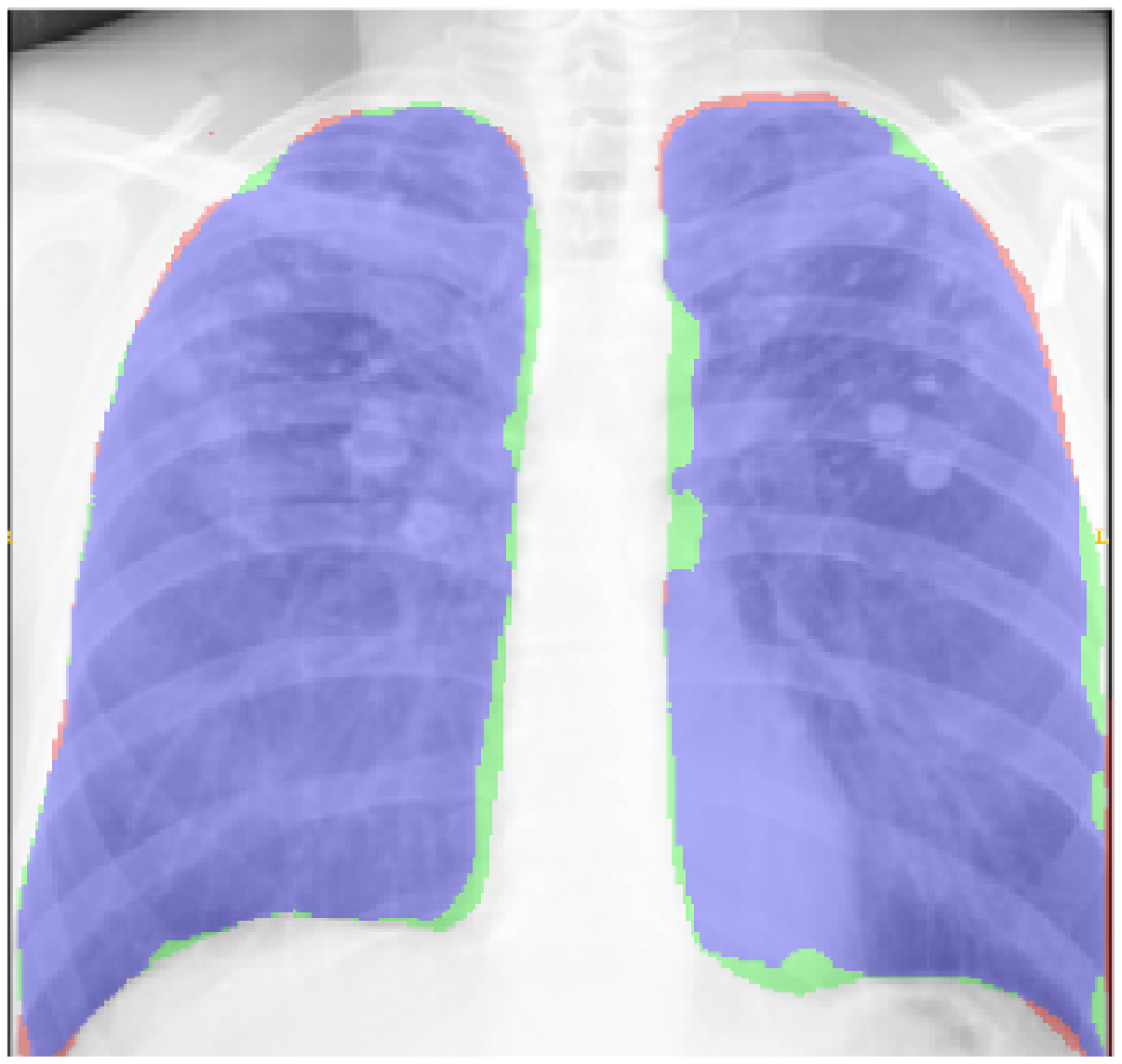}
\caption{\footnotesize{DSC=0.968}}
\end{subfigure}
\caption{\footnotesize{{\color{black}Qualitative lung field segmentation results obtained using the state-of-the-art U-Net-based architecture proposed by \cite{wang2017segmentation} for the segmentation of structures in CXR. The cases shown were randomly chosen from the dataset. The segmentation labels obtained are overlaid in over the input CXR. The blue region denotes the overlap area between the ground expert segmented manual ground truth and the segmentation produced using \cite{wang2017segmentation}, the green region denotes ground truth area, and the red region denotes the segmentation obtained using U-Net. As can be seen from the overlaid labels that shape specificity is not preserved by the U-Net architecture.}}}
\label{fig:qualitativeUnet}
\end{figure}

{\color{black}We compared the segmentation performance obtained through our approach to the results reported by the state-of-art methods using three widely used metrics (overlap, average contour distance (ACD), and DSC) in Table \ref{table:methodsPerformance}. The table reports the performance on both lungs. The results reported here for our method are obtained on the original $2048\times 2048$ images and not the down-sampled version. None of the other methods includes the retro-cardiac region within the segmentation. In addition, we also compared the segmentation performance with the U-Net based architecture proposed by \cite{wang2017segmentation}: the current state-of-the-art convolutional neural network for biomedical image segmentation. The U-Net architecture and its derivatives have been extensively used for segmentation in radiological and histological images, providing some of the most accurate and satisfactory performances \cite{kamnitsas2017efficient, sirinukunwattana2016locality, cciccek20163d, chen2016dcan}. The U-net architecture is a fully convolutional network, which includes shortcut connections between a contracting encoder and a successive expanding decoder. The quantitative overall segmentation performance as well as on individual datasets (JSRT, BTP, CNHS) using  Wang et al.'s approach is also reported in Table \ref{table:methodsPerformance} for segmentation labels with and without retro-cardiac space. {\color{black}Exact same architecture and hyper parameters as reported in Wang's paper were used except the post-processing step which was omitted for fair comparison since our proposed approach does not use any post-processing. Although, a range of hyper-parameters were tested; however, ones proposed by Wang et al. were found to be optimal for the task.}

Fig. \ref{fig:qualitative} presents the qualitative results of performing the lung segmentation using the proposed pipeline (ESL+MaShDL). The figure provides a visual insight on how inclusion of retro-cardiac region results in the segmentation label that is independent to the shape and structural changes in the close-by anatomical structures such as heart. For comparison purposes, similar qualitative results for the lung field labels obtained using the method proposed in \cite{wang2017segmentation} are provided in Fig. \ref{fig:qualitativeUnet}. As predicted before, the shape specificity is not preserved for the lung field labels obtained using \cite{wang2017segmentation}. This is further evident through the results presented in Table \ref{table:methodsPerformance}. Moreover, unlike the proposed method, the U-net architecture uses an overlapping-based objective function (e.g., cross-entropy) which provides satisfactory results in cases with reduced shape variability. However, in the particular case of thoracic radiographs, the lung field labels without retro-cardiac space present higher shape variability than those observed when including this region. This could be a possible explanation of a slightly better overlapping-based performance (i.e., Overlap and DSC) by U-Net \cite{wang2017segmentation} when including the retro-cardiac space than without including it.}

\section{Discussion and Conclusion}
This work introduced a generic representation learning framework for the deformable object segmentation via space (translation, orientation, anisotropic scaling) and shape parameter estimation. The boundary detectors in the conventional statistical shape models (SSM) do not work consistently well on the data with complex patterns or with poor contrast and edge information. Furthermore, since the SSMs are known to be sensitive to initial shape estimation, an efficient learning-based mechanism to estimate the space parameters (translation, scale, and orientation) was also presented in this work to initialize the mean shape. Our solution to space parameter learning, ensemble space learning (ESL), was significantly more accurate than current state-of-the-art marginal space learning (MSL) \cite{zheng2008four} and marginal space deep-learning (MSDL) \cite{ghesu2016marginal} approaches as demonstrated through rigorous experiments. Although ESL has the potential to be generically applicable for the localization of objects of interest in 2D/3D images; however, the algorithm, in its current form, assumes symmetry (such as lung field) of the object of interest as well as the neighborhood context information for efficiency purposes. Therefore, while ESL is envisioned to demonstrate best performance for the localization of objects in medical images where organ symmetry as well as the contextual information can be somewhat guaranteed; for the general computer vision tasks, the ESL may need to be modified for optimal results. Furthermore, due to the existence of clinical acquisition protocols, large rotational variation is not expected in various CXR images; therefore, rotation estimation is still performed sequentially after translation and scaling rather than independently. The method can be easily modified for tasks where large rotational variation in the training data is expected. 

Furthermore, for a given performance accuracy, our formulation for marginal shape deep learning (MaShDL) estimated deformable shape parameters significantly faster than the conventional SSM-based methods. As has been stressed throughout in the manuscript, MaShDL extends the ASM into deep learning realm; which results in better overall accuracy as demonstrated by rigorous set of experiments. However, since the mathematical framework behind MaShDL is still similar to the ASM, some of the limitations of the traditional ASM that are part of the mathematical framework exists in the MaShDL framework as well, that includes: (1) the tedious task of labeling training images that becomes unacceptable especially with the large training set. As pointed out previously in the manuscript that although the current scheme of six manually defined landmarks was found sufficient for accurate lung field segmentation (Dice score between the manual ground truth label and the label obtained using the interpolated landmarks  $=0.994\pm 0.001$), different number of manually annotated landmarks can be tested based on the application and the object of interest. (2) Similar to the traditional ASM, parameters such as the number of modes of variation still need to be specified. Furthermore, as the difference between the positive and negative hypothesis becomes more subtle at higher modes; in order to learn modes beyond a certain limit, approaches such as the use of deeper networks and mode-dependent thresholds for hypotheses testing need to be investigated. (3) The use of global statistical shape models by approaches like ASM is one of the most successful method to impose shape and anatomical constraints in medical image segmentation. However, while providing robust and anatomically accurate constraints, it also limits the flexibility of the method to deal with small localized shape details, such as the region around the diaphragm in chest radiographs. To overcome this limitation, we intend to extend our previous work on partitioned shape modeling \cite{mansoor2016deep} to MaShDL in the future. (4) For the specific application of lung-field segmentation, certain extreme cases of scoliosis that the mean shape model failed to capture accurately during training may show suboptimal accuracy. Although we exemplified an application of our framework through the segmentation of the lung field from CXR using diversified populations (i.e., age, pathology, and source); however, even with this diversity, as long as the standard acquisition protocols of routine clinical environment were followed, the algorithm was designed to robustly handle variation in shape. Our framework is applicable to general deformable object segmentation in both 2D and 3D image data, as a faster and potentially more accurate alternative to statistical appearance and shape model. 

\bibliographystyle{IEEEtran}
\begin{footnotesize}
\bibliography{IEEELung}
\end{footnotesize}
\end{document}